%% file: main.tex
\documentclass[10pt]{article} 
\usepackage{tmlr}
\usepackage{microtype}      
\usepackage{graphicx}       
\usepackage{subfigure}      
\usepackage{booktabs}       
\usepackage{hyperref}       
\usepackage{subcaption}
\usepackage{wrapfig}
\usepackage{amsmath}
\usepackage{amssymb}
\usepackage{mathtools}
\usepackage{amsthm}
\usepackage{enumitem}
\usepackage{float}
\usepackage[capitalize,noabbrev]{cleveref}
\usepackage{algorithm}
\let\AND\relax
\usepackage{algorithmic}
\usepackage{xcolor}
\usepackage{array,multirow}
\usepackage{url}
\theoremstyle{plain}

\newtheorem{example}{Example}

\theoremstyle{definition}

\theoremstyle{remark}

\usepackage{xcolor} 

\newcommand{\reb}[1]{\textcolor{black}{#1}}
\newcommand{\Hquad}{\hspace{0.4em}}

\input{math_commands.tex}

\usepackage{hyperref}
\usepackage{url}

\title{The Over-Certainty Phenomenon in Modern Test-Time \\Adaptation Algorithms}

\author{\name Fin Amin \email samin2@ncsu.edu \\
      \addr Department of Electrical and Computer Engineering \\
      North Carolina State University
      \AND
      \name Jung-Eun Kim\thanks{Corresponding author} \email jung-eun.kim@ncsu.edu \\
      \addr Department of Computer Science \\
      North Carolina State University}

\def\month{07}  
\def\year{2025} 
\def\openreview{\url{https://openreview.net/pdf?id=AGQRij8iUC}}

\begin{document}

\maketitle

\begin{abstract}
When neural networks are confronted with unfamiliar data that deviate from their training set, this signifies a domain shift. While these networks output predictions on their inputs, they typically fail to account for their level of familiarity with these novel observations. Prevailing works navigate test-time adaptation with the goal of curtailing model entropy, yet they unintentionally produce models that struggle with sub-optimal calibration—a dilemma we term the over-certainty phenomenon. This over-certainty in predictions can be particularly dangerous in the setting of domain shifts, as it may lead to misplaced trust. In this paper, we propose a solution that not only maintains accuracy but also addresses calibration by mitigating the over-certainty phenomenon. \reb{To do this, we introduce a certainty regularizer that dynamically adjusts pseudo-label confidence by accounting for both backbone entropy and logit norm.} Our method achieves state-of-the-art performance in terms of Expected Calibration Error and Negative Log Likelihood, all while maintaining parity in accuracy.

\end{abstract}

\input{Introduction}

\input{relatedWork}

\input{observations}

\input{exp}

\input{results}

\input{conclusion}

\bibliography{main}
\bibliographystyle{tmlr}

\appendix

\input{appendix}

\end{document}

%% file: math_commands.tex
\usepackage{amsmath,amsfonts,bm}

\def\eqref#1{equation~\ref{#1}}

\def\1{\bm{1}}

\DeclareMathAlphabet{\mathsfit}{\encodingdefault}{\sfdefault}{m}{sl}
\SetMathAlphabet{\mathsfit}{bold}{\encodingdefault}{\sfdefault}{bx}{n}



%% file: Introduction.tex



\section{Introduction}

When encountering new environments, humans naturally adopt a cautious approach, assimilating the novelty to guide their decision-making. This inherent ability to assess unfamiliarity and adjust certainty has not been entirely emulated in artificial neural networks. Unlike humans who might exhibit hesitation in unknown situations, many test-time adaptation (TTA) algorithms lack an explicit mechanism to modulate certainty in response to the novelty or unfamiliarity of their inputs.

Deep learning has never been a stranger to the challenges of uncertainty. Over the past few years, the miscalibration problem of modern neural networks has gained substantial attention, as highlighted by works such as \cite{guo2017calibration}, \cite{abdar}, \cite{liang2017enhancing}, and \cite{pampari2020unsupervised}. However, how TTA algorithms themselves alter calibration is understudied. In this paper, we uncover the \textit{over-certainty phenomenon}, a phenomenon that plagues many modern TTA algorithms by harming model calibration.

\begin{figure}[h] 
  \centering
  \includegraphics[width=0.65\linewidth]{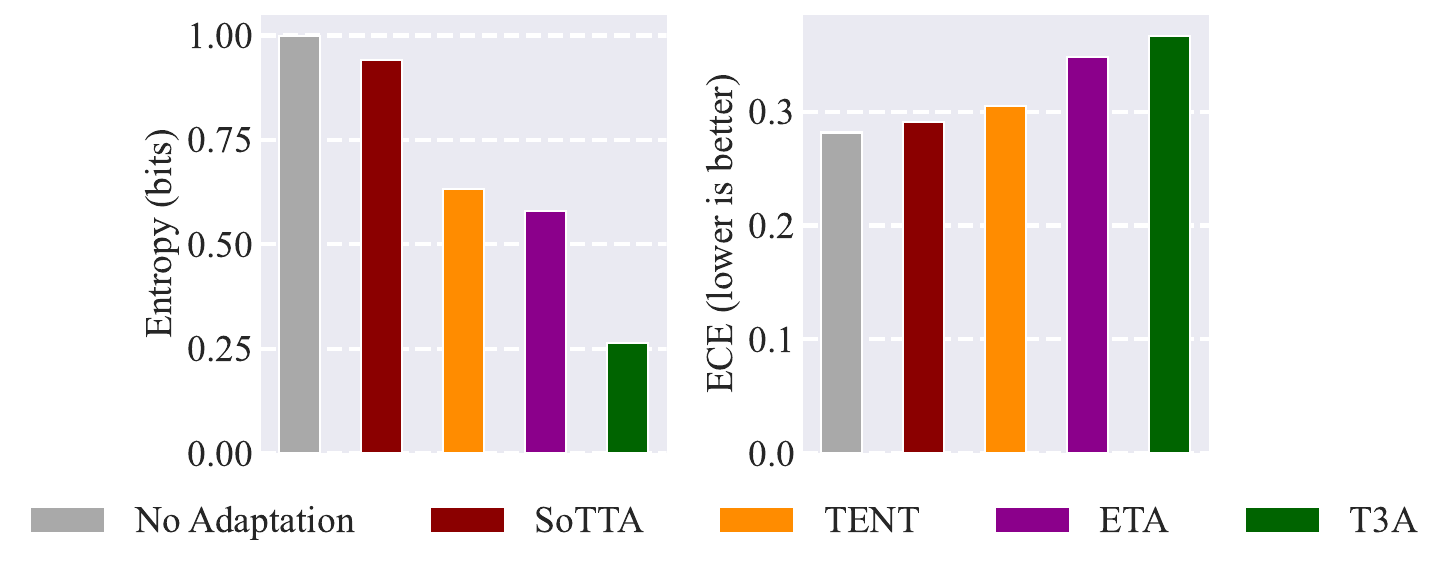}
  \caption{This chart shows four modern TTA algorithms adapting EfficientNet to the \textit{clipart} domain. Minimizing entropy is a common objective in recent work. However, this can have consequences on model calibration. }
  \label{fig:lowEntropy_badEce}
\end{figure}

A prevailing strategy among TTA algorithms is the minimization of entropy, either as an explicit target or as an inherent by-product of their methodology. And while this might bolster accuracy metrics, our research indicates a concerning trend: excessive entropy reduction can be detrimental to model calibration. What makes this trend more problematic is that it occurs within the context of a new domain, where epistemic uncertainty should typically be greater. {In our work, we measure certainty as the inverse of Shannon entropy} of a model's output after a softmax operation: $ H(f(x))^{-1} = 
 (\mathtt{Entropy_2}(\mathtt{SoftMax}(f(x))))^{-1}$.

To further frame our discussion, TTA is used when a model, trained on a source domain $(X_s, Y_s)$, is presented with the challenges of a different yet analogous target domain $(X_t)$ with no labels. {In our problem setting, we do not assume access to $X_t$ until we adapt.} The nuances between these domains, commonly termed as domain shift, can introduce significant disruptions in model performance. TTA, in its essence, aspires to adapt the insights harvested from the source domain and apply them proficiently to the target domain, bypassing the need for labeled data in the latter. \reb{TTA is difficult because models must generalize to unfamiliar distributions without access to labeled data or strong assumptions about how the target domain relates to the source \cite{is_ood_learnable,impossibility_theorems_for_uda}.} Furthermore, as we will elaborate in the next section--it can be especially problematic to be too certain within the context of a domain shift.

With the purpose of addressing these intertwined challenges, we introduce \textit{Dynamic Entropy Control}. This TTA technique seeks to augment accuracy and improve model calibration. By interweaving calibration into the core learning process, we produce a TTA adjustment algorithm that jointly improves accuracy while managing epistemic uncertainty. To summarize, our contributions are:

\begin{itemize}
    \item The identification of the over-certainty phenomenon. We emphasize that our work is not about the well-studied tendency of models to become less calibrated under distribution shift \cite{pampari2020unsupervised, canyoutrust}. Instead, we show how a recent trend in the design of TTA algorithms can yield models that are even more miscalibrated than the unadapted baseline. To the best of our knowledge, no prior work—including \cite{entropyEnigma}—explicitly analyzes this pattern while also providing thorough empirical evidence and mechanistic insight into why over-certainty emerges during adaptation.

    \item \emph{Dynamic Entropy Control}, a new TTA algorithm that achieves SOTA calibration in all of the four datasets and SOTA accuracy uplifts in the majority of domain shifts. 

\end{itemize}

Our study is scoped to {test-time adaptation algorithms that follow the prevailing trend of minimizing entropy on unlabeled observations} in the classification setting. While we acknowledge the existence of TTA methods that do not follow this paradigm (e.g.,~\cite{shot,srdc}), they fall outside the focus of our analysis. \textbf{Unless otherwise stated, all mentions of ``TTA algorithms'' in this work refer specifically to entropy-reduction-based approaches.}

\subsection{Why Care About Over-Certainty?}\label{sec:why_care_about_ocp}

\noindent We emphasize that calibrating uncertainty estimates has ramifications that go well beyond quantitative metrics. For instance, in real-world scenarios that rely on trustworthy estimates of model confidence (e.g., autonomous driving, active learning, anomaly detection), an overconfident model can lead to harmful misjudgments \cite{UDA_autonomousDriving}. As a concrete illustration, consider a self-driving car that uses a classifier to detect whether a neighboring lane is occupied: 

\begin{itemize} 
\item \emph{Pre-adaptation (low certainty):} The vehicle’s classifier might cautiously predict ``safe to switch lanes'' but at a low confidence. This low certainty prevents the car from making a risky maneuver. 

\item \emph{Post-adaptation (artificially high certainty):} After applying a traditional TTA algorithm, the same classifier might become \emph{overly certain} about its ``safe to switch lanes'' prediction. In reality, if this confidence is misplaced, the system can make a high-stakes error that jeopardizes safety. 
\end{itemize} 

Thus, our approach seeks to preserve or improve classification accuracy \emph{and} produce calibrated uncertainty estimates, ensuring that decisions made based on confidence thresholds are more reliable. Such reliability is also valuable in other domains that hinge on calibrated outputs, such as fairness-sensitive applications \cite{ multicalibration, environmentInferenceInvariant}, where miscalibration can exacerbate biases or distort outcome distributions.

\noindent \textbf{Fairness Considerations.}
As suggested in \cite{onFairnessAndCalibration}, there is a growing recognition of how calibration can influence fairness. Overconfident models may disadvantage certain groups when high-confidence predictions guide resource allocation or risk assessments. Ensuring well-calibrated probabilities is therefore crucial but often overlooked in current TTA approaches.


%% file: relatedWork.tex
\section{Related Work}

 Our literature survey covers methodologies catered towards updating a neural network on unlabeled data. For the sake of brevity, we will refer to unlabeled data as ``observations.'' This section gives an overview of the work done to improve networks on the fly. We start by introducing earlier work, such as dictionary learning techniques and lead our way into recent developments. We elaborate the algorithms we compare with in detail to give mechanistic insight on why over-certainty happens. The next section covers how calibration is measured and improved in the context of a domain shift. Lastly, we discuss assessing the reliability of an observation. Furthermore, we provide a supplementary discussion of related work in Appendix~\ref{sec:appendix_further_discussion_related_work}.

 \subsection{The Test-Time Entropy Reduction Paradigm}

 The phrase ``self-taught learning'' was coined by \cite{raina2007self}. In this work, the authors utilize observations to find an optimal sparse representation of said observations. This sparse representation is used to train their model in lieu of the ordinary training set to improve out-of-distribution (OOD) performance.

 
  Work in this field has extended to a variety of approaches such as the use of pseudo-labeling to exploit the existing model's predictions as target labels \cite{psuedo_label,sourceSpecificNets,uda_survey}. Pseudo-labeling can be thought of under the guise of knowledge distillation (KD) \cite{hinton_distil,kd_survey}. KD is a transfer learning paradigm where a large neural network, known as the teacher, transfers ``knowledge'' to a smaller ``student'' network. Succinctly, the student is trained to match the output of the teacher when given the same input as the teacher \cite{does_knowledge_distillation_work}. In pseudo-labeling, the teacher and student are the same network.

 The TENT algorithm introduces the trend of \textbf{test-time entropy minimization} \cite{TENT}. Entropy minimization can be thought of as using pseudo labels with the cross entropy loss function. In other words, the entropy minimization works by using gradient descent to minimize:
\begin{align}
    L_{TENT} &= -\sum_{y \in C} f(y|x) \log f(y|x) \label{eq:tent_loss}
\end{align}
to update the model's batch-normalization parameters. 

More recent advancements in this trend include EATA/ETA \cite{eata}, T3A \cite{t3a} and SoTTA \cite{sotta}. ETA\footnote{The authors of EATA/ETA introduce two similar algorithms, for our paper, we focus on ETA which performs the best between the two.} advances on TENT by making sure that observations are \textit{reliable} and \textit{non-redundant} before they are used for updating the batch-normalization parameters. To do this, they compute a sample adaptive weight, $\mathcal{S}(x)$, for each observation before minimizing entropy:
 \begin{align}
    L_{ETA} &= -\mathcal{S}(x)\sum_{y \in C} f(y|x) \log f(y|x) \label{eq:eta_loss} 
\end{align}
where $\mathcal{S}(x)$ is a function of the entropy of the model towards the batch sample (i.e., the reliability) and the similarity to what it has seen before (i.e., non-redundancy). Similar to the aforementioned methods, SoTTA minimizes entropy via sharpness-aware-minimization (SAM) \cite{SAM}. The algorithm employs high-confidence uniform sampling to create a memory bank of size $N_{SoTTA}$ which stores reliable and class-balanced observations. This is done by using confidence to asses if an observation should be used for adaptation. Confidence is defined as:
\begin{align}
    \mathcal{C}_{f}(x) &= \max_{i=1,\ldots,n} \frac{\exp(f(x)_i)}{\sum_{j=1}^{n} \exp(f(x)_j)} \label{eq:confidence}
\end{align}

If $\mathcal{C}_{f}(x) > \mathcal{C}_{0}$, where $\mathcal{C}_{0}$ is some pre-defined confidence threshold, then $x$ is saved into memory. Afterwards, the SAM optimizer is used to minimize entropy (equation~\ref{eq:tent_loss}) via {two} backpropagation steps. The T3A algorithm \cite{t3a} differs from the previous three as it focuses on updating the \textit{prototypes} \cite{prototype_learning} of each class during test time:
\begin{align}
\small
 S_k^t = \begin{cases}
    S_k^{t-1} \cup \{ f(x) \}, & \text{if } \hat{y} = y_k \\
    S_k^{t-1}, & \text{else}. \label{eq:t3a_filter}
        \end{cases} \\
    c_k = \frac{1}{{\lvert S_k \rvert}} \sum_{z \in S_k} z 
\end{align}
where $c_k$ represents the centroid of the prototypes of a class $k \in C$, where $C$ is the number of classes. We define the feature extractor, $\psi$, as all the layers of the backbone before the final dense layer. The final dense layer, $\phi$, is what we refer to as the classifier, it is composed of the class centroids. We denote the output of the feature extractor as $z = \psi(x)$.


 Unlike TENT, SoTTA, or ETA, T3A does not explicitly reduce entropy as it does not use a loss function, however the authors claim that entropy reduction is an effect of using their algorithm. Similar to ETA, this algorithm filters less reliable samples during equation \ref{eq:t3a_filter} by only keeping the $M$ lowest entropy prototypes for each class. Therefore, the algorithm stores $C\cdot M$ prototypes.


 \subsection{Neural Network Calibration} \label{calibration and Uncertainty survey}


Neural network calibration has been of intense interest in recent years due to the critical role of confidence values, which reflect the probability assigned to predictions, in various applications. For instance, BranchyNet \cite{branchy} uses neural network confidence to enable early exits for faster inference, relying on high confidence at intermediate layers. However, \cite{deepprobability} highlights the prevalent issue of certainty calibration in deep networks, where models often display overconfidence or underconfidence, likely due to overfitting during training. \cite{canyoutrust, revisitingCalibration} explore this concept further by measuring a model's Expected Calibration Error (ECE) and Negative Log-Likelihood (NLL). ECE measures how closely the confidence levels of a model's predictions match the actual probability of those predictions being correct. It calculates the average absolute difference between predicted confidence and the true outcome frequencies, providing a metric for the reliability of the model’s probabilistic outputs. NLL, on the other hand, captures both the model's calibration and sharpness by quantifying how well the predicted probabilities align with the observed outcomes, penalizing overconfident yet incorrect predictions more heavily. {We follow their convention and use  ECE and NLL as our measures for calibration error.} They notice models calibrated on the validation set tend to be well calibrated on the test set, but are not properly calibrated to shifted data. 

Recent work has also investigated solutions to this phenomenon. \cite{guo2017calibration} discusses a technique known as temperature scaling while \cite{logit_norm} approaches this problem by regularizing the logit norm. More classical solutions to this problem exist as well; \cite{delvingdeepLabelSmoothing} and \cite{when_does_label_smooth_help} consider label smoothing to address this issue. Note that these techniques are addressed at calibrating the underlying backbone but have \textit{not} been investigated with respect to TTA algorithms themselves. 

\subsection{Detecting Out of Distribution Data and Assessing Reliability} \label{sec:detect_ood}


An increasing body of research examines how to assess whether and how closely a new observation aligns with a model’s training distribution. For example, the authors of \cite{learning_competitive_descrimintive_re} observe that if an autoencoder was trained to reconstruct inliers, it would have a greater reconstruction error when reconstructing OOD data. \cite{AnoGAN} and \cite{effecient_gan_based_reconstruc} approach this issue by observing that the discriminator of a GAN learns whether or not a given input is an inlier. Many other works delve into this domain \cite{openMax,vl_rep,understand_Feature_norm,ood_via_prior,meta_ood_detect,long_tail_ood, is_ood_learnable}. Regarding the TTA algorithms we compare against, the most common proxy for reliability is entropy on the observation.


%% file: observations.tex
\reb{\section{What is the Over-Certainty Phenomenon?}}
\label{sec:over-certain}

In this work, we present evidence for what we dub the \textit{over-certainty phenomenon} (OCP) of contemporary TTA approaches. This phenomenon is that TTA algorithms tend to miscalibrate their underlying backbone networks by causing their predictions to be excessively certain. Modern TTA algorithms often strive to decrease test-time entropy. However, as shown in Fig.~\ref{fig:lowEntropy_badEce}, this entropy reduction may increase ECE and NLL because the models become overly certain on their predictions. 

This phenomenon of existing algorithms causing models to become overly certain presents itself across many other datasets. A compelling example is given in Table~\ref{tab:TIN_C_metrics} which agglomerates calibration errors over 15 domain shifts; in this table, we see a clear trend of entropy reduction (certainty increasing) and sub-optimal calibration. Another example is provided in Table~\ref{tab:HO_metrics}, T3A reduces entropy by a factor of about 4 in the \textit{art, clipart} and \textit{product} domains. As before, it causes ECE to worsen compared to the baseline. We do not claim that TTA algorithms should \emph{always} strive to increase backbone uncertainty; poor calibration can also be caused by under-certainty and there exist cases where reducing entropy compared to baseline improves calibration. However, we find that the resulting calibration is still sub-optimal. Despite these complexities, our investigation reveals a consistent pattern: \emph{the over-certainty phenomenon causes sub-optimal model calibration}, a significant concern for safety, robustness, and reliability.

\input{tables_charts/HO_Table_mobileNet}
\subsection{What Causes the Over-Certainty Phenomenon?}

We identify two plausible causes of the OCP, the first issue is that modern TTA algorithms aim at minimizing backbone entropy too aggressively. In the case of TENT, ETA, and SoTTA, their loss functions, \eqref{eq:tent_loss} and \eqref{eq:eta_loss}, explicitly aim at reducing a model's entropy. Regarding TENT, there is no regularization of this process. In the case of ETA, the algorithm uses a \textit{reliability score}, $S(x)$, which aims at weighing observations differently but does not regularize the distributions of the pseudo-labels. Unlike the other two, SoTTA minimizes entropy twice per iteration. The authors of T3A claim that entropy reduction is an effect of using their algorithm. In fact, they show in certain datasets T3A reduces entropy more than TENT does. 

Another issue is how existing methods evaluate observation \textit{reliability}, the suitability of a model's prediction for use for adaptation. Previous works, ETA and T3A, tap into the power of model certainty, using it to weigh the influence of observations. ETA assesses reliability by ensuring that observations meet a certain entropy threshold; similarly, T3A uses entropy to sort the importance of class prototypes. However, as prior work has shown, there are drawbacks in using entropy as a proxy for reliability in this manner \cite{logit_norm}. To illustrate our point, we give a toy example of how using entropy can lead to a misleading conclusion:

\begin{example} \label{example: no logit norm}
Suppose that we analyze the classifier while classifying between two classes with class centroids, $c_0$ and $c_1$. This is done by taking the output of the feature extractor, $\psi(x) = z$, and computing the dot product between the centroids and $z$. 
\begin{align}
    g = [z\cdot c_0, z\cdot c_1] 
\end{align}

Consider \( g_{t1}, g_{t2} \) and \( g_s \) as vectors representing the inner products related to two observations, $x_{t1}$ and $x_{t2}$, and to a specific training sample, $x_{s}$. Specifically, $g$ corresponds to the dot products between the output of the feature extractor and the class centroids. As an example, let's assume: 
\begin{align*}
    g_s = [8.0, 7.29]; \Hquad g_{t1} = [1.92, 1.00]; \Hquad g_{t2} = [6.10, 6.50];
\end{align*}
If we take the softmax of these vectors and compute the entropy, we get $\mathtt{Entropy_2}(\mathtt{SoftMax}(g))$ for $g_s, g_{t1}$ and $g_{t2}$, as 0.92 bits,  0.86 bits and 0.97 bits, respectively.

\end{example}

\begin{figure}[!ht]
  \centering

  \begin{minipage}[t]{0.46\linewidth}
    \centering
    \includegraphics[width=0.85\linewidth]{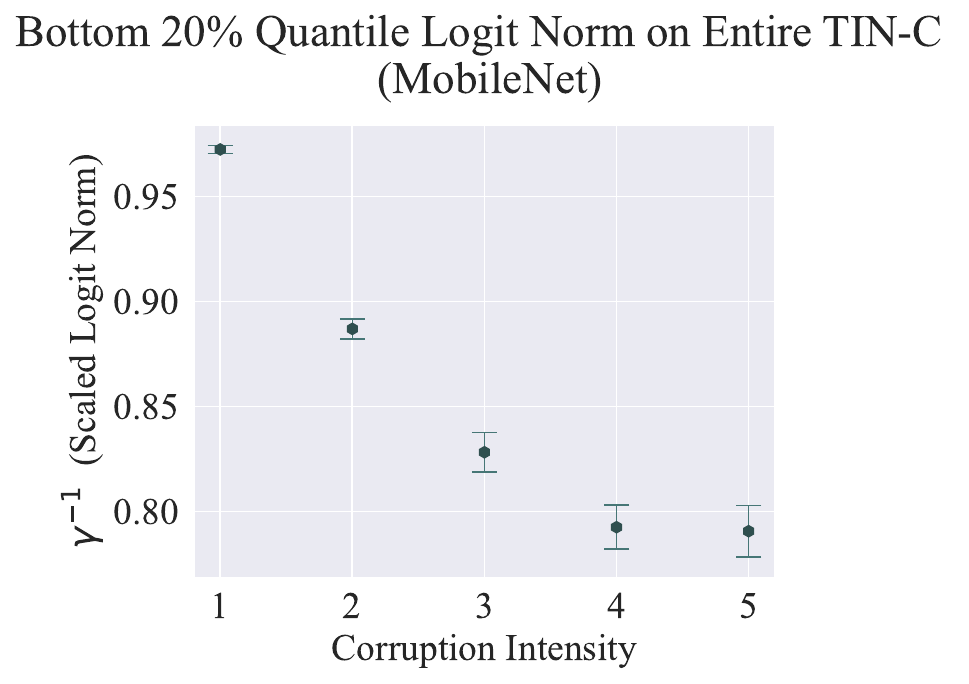}
    \captionof{figure}{As domain shift intensity increases, the bottom quantile of the observation logit norm decreases. The $\gamma^{-1}$ value (step 8) from our $\texttt{CCR}$ algorithm represents the ratio between the $l_2$ norms of the observation logits, $z$, and the training-set logits, $\kappa$. As $z$ decreases, $\gamma$ increases; this regularizes low-logit-norm observations more aggressively (step 9). Vertical bars indicate domain-to-domain standard deviations.}
    \label{fig:logitNorm}
  \end{minipage}
  \hfill
  \begin{minipage}[t]{0.48\linewidth}
    \centering
    \includegraphics[width=\linewidth]{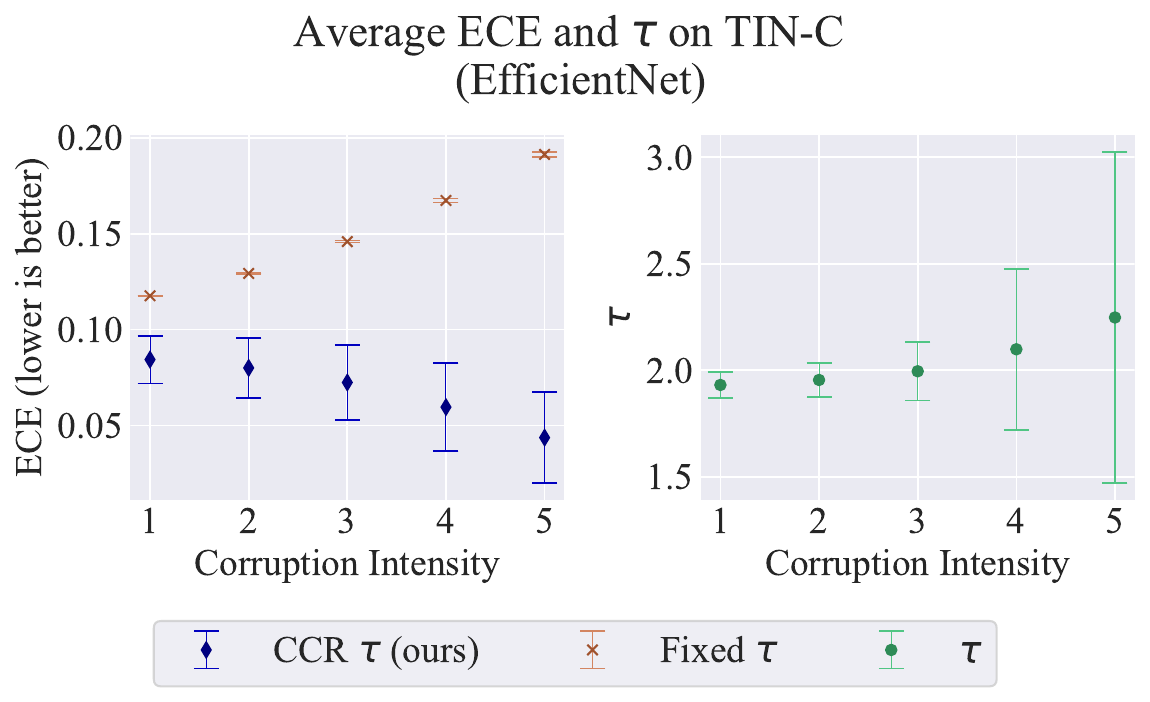}
    \captionof{figure}{Our ablation study highlights the effectiveness of the $\texttt{CCR}$ algorithm compared to a fixed $\tau$ optimized for minimal ECE on the source-domain training set. The rightmost figure displays the computed $\tau$ values, with vertical bars indicating domain-to-domain standard deviations.}
    \label{fig:ablation}
  \end{minipage}

\end{figure}

Notice that if we consider the entropy of these three vectors as a proxy for reliability, we would consider $x_{t1}$ to be more reliable than $x_{t2}$, despite $x_{t2}$ having considerably greater inner product with the class centroids. It is highly likely that the values of $g_{t1}$ occurred due to spurious feature correlations between $x_{t1}$ and the class centroids. In fact, in the scenario above, $x_{t1}$ would be deemed to be more reliable than the genuine source domain observation $x_s$. Note that an analogous remark could be made on using confidence instead of entropy. Our analysis is not contrived; in Fig.~\ref{fig:logitNorm}, as we increase the domain shift intensity, the observation logit norm decays and has higher variance.

Furthermore, existing approaches do not consider a model's certainty on the source domain. For example, ETA's reliability paradigm rejects observations which have $H \geq 0.4 \cdot \ln{(C)}$. What if we \textit{expect} (i.e. there was high entropy on the training set) the model to have high entropy?  By only evaluating the target domain's certainty without juxtaposing it against the source domain certainty, there is a lack of reference in terms of assessing the reliability of the observation. We address this issue via the $h_0$ parameter explained in the next section.  


\section{The Dynamic Entropy Control TTA Algorithm\label{cd_algo}}

To ameliorate the over-certainty phenomenon, we introduce Dynamic Entropy Control (\texttt{DEC}) (Algorithm \ref{alg:cd_actual_algo}). DEC refines the model's certainty levels, aligning them more closely with its actual accuracy, by selectively adjusting the temperature parameter during the pseudo-labeling process (line 6). This is achieved without directly altering ground truth labels, instead focusing on the tempering of logits through temperature adjustments. {Our approach can be implemented using a single model which alternates weights. To facilitate understanding, we explain our algorithm as if it uses two distinct models}: the teacher and the student. The ``teacher'' model, $f_{te}$, is simply our backbone before any adaptation. The ``student'' model, $f_{s}$, is the model that is adapted.

\noindent
\begin{minipage}[t]{0.48\textwidth}
\begin{algorithm}[H] 
\small
\caption{$\texttt{Compute Certainty Regularizer}$ \\$\texttt{(CCR)}$}
\label{alg:compute_T}
\textbf{Input}: $f_{te}(X), h_0, h_{max}, t_{min}, t_{max}, \kappa$\\
\textbf{Output}: $T_{vec}$
\begin{algorithmic}
\STATE $Z = f_{te}(X)$ \COMMENT{get logits}
\STATE $H_{vec} = \mathtt{Entropy_2}(\mathtt{SoftMax}(Z))$ \COMMENT{entropy for each sample}
\STATE $H_{diff} = H_{vec} - h_0 $ \COMMENT{compare entropy with source entropy}
\STATE $H_{scaled} = \mathtt{sigmoid}(H_{diff}/ \sqrt{h_{max}})$ \COMMENT{scale between [0,1]}
\STATE Init. $T_{vec}$
\FOR{$h_i \in H_{scaled}$ and $z_i \in Z$}
\STATE $t_i = t_{min} + h_i \cdot (t_{max} - t_{min})$ \label{eq:ti}
\STATE $\gamma_i = \frac{\kappa}{\|z_i\|_2}$ \COMMENT{Scale the logit norm} \label{eq:gammai}
\STATE $\tau_i = \gamma_i \cdot t_i$ \COMMENT{Adjust regularizer via logit norm} \label{eq:taui}
\STATE Store $T_{vec} \leftarrow \tau_i$
\ENDFOR
\STATE \textbf{return} $T_{vec}$
\end{algorithmic}
\end{algorithm}
\end{minipage}
\hfill
\begin{minipage}[t]{0.48\textwidth}
\begin{algorithm}[H] 
\small
\caption{\texttt{{Dynamic Entropy Control}} \\ \texttt{(DEC)}}
\label{alg:cd_actual_algo}
\textbf{Input}: $f_{te}, f_s, h_0, X, \kappa$\\
\textbf{Parameters}: $t_{min}, t_{max}, h_{max},  \lambda$\\
\textbf{Output}: $f_s^+$
\begin{algorithmic}
\STATE $T_{vec} = \mathtt{CCR}(f_{te}(X), h_0, \kappa, h_{max}, t_{min}, t_{max})$
\STATE Init $loss$
\FOR{$x_i \in X$ and $\tau_i \in T_{vec}$} 
\STATE $s_{si} = \text{\textcolor{black}{\texttt{SoftMax}}}(f_s(x_i))$
\STATE $s_{ti} = \text{\textcolor{black}{\texttt{SoftMax}}}\left(\frac{f_{te}(x_i)}{\tau_i}\right)$ \COMMENT{smoothen teacher labels}
\STATE $l = (\mathtt{avg}(T_{vec}))^2 \cdot \mathtt{CE}(s_{si}, s_{ti})$
\STATE Store $loss \leftarrow l$
\ENDFOR
\STATE $L = \mathtt{avg}(loss)$
\STATE $f_s' \leftarrow \theta_s - \lambda \nabla L(\theta_s)$ 
\STATE $f_s^+ = \mathtt{Temperature\_Scale}(f_s', \mathtt{avg}(T_{vec}))$
\STATE \textbf{return} $f_s^+$
\end{algorithmic}
\end{algorithm}
\end{minipage}
\vspace{1.0em}

Dynamic Entropy Control calibrates model predictions under distribution shift by adaptively controlling the sharpness of pseudo-labels. Instead of applying a fixed or global temperature, DEC adjusts the certainty of each pseudo-label using a scalar we call the certainty regularizer. This scalar is computed per observation, based on how uncertain the model appears on that input--taking into account both the predicted entropy and the norm of the logit vector. This mechanism allows DEC to smooth unreliable predictions more aggressively, helping the model avoid overconfident errors during test-time adaptation.

\reb{\subsection{The Certainty Regularizer}\label{sec:certainty regularizer}}

\reb{The key component of DEC is the certainty regularizer, $\tau_i$, a per-sample temperature scalar returned by the $\texttt{Compute Certainty Regularizer}$ routine (Algorithm~\ref{alg:compute_T}). This scalar adjusts how confidently the model treats each pseudo-label. To compute $\tau_i$, we first compare the model's entropy on an input $x_i$ to its typical behavior on the source domain. This reference entropy, $h_0 = \mathbb{E}[\mathtt{Entropy_2}(\mathtt{SoftMax}(f_{te}(X_s)))]$, reflects the average certainty of the teacher model before adaptation. The difference $H(x_i) - h_0$ is then passed through a sigmoid that is scaled using $h_{\max} = \log_2(C)$, the maximum possible entropy over $C$ classes. This yields a normalized “entropy contrast” in $[0,1]$, which is mapped to a temperature range defined by $t_{min}$ and $t_{max}$. These bounds control the minimum and maximum smoothing applied to any sample.}

\reb{To further refine this adjustment, DEC incorporates logit-norm awareness by modulating the entropy-derived temperature with the inverse norm of the model's logits. Specifically, for each sample, we compute the $\ell_2$ norm of its logit vector $\mathbf{z}_i = f_{te}(x_i)$ and scale the temperature accordingly. The scaling factor is given by $\gamma_i = \kappa / |\mathbf{z}_i|_2$, where $\kappa$ is the median logit norm computed over the source-domain samples. This has the effect of increasing $\tau_i$ when the model’s confidence is low in logit space, thereby injecting more smoothing into uncertain predictions. The final certainty regularizer is then given by $\tau_i = t_i \cdot \gamma_i$, combining both entropy contrast and logit-norm awareness to calibrate confidence per sample:}

\[
\tau_i = (\underbrace{t_{\min} + \mathtt{sigmoid}\left( \frac{H(x_i) - h_0}{\sqrt{h_{\max}}} \right)(t_{\max} - t_{\min}))}_{\text{entropy contrast }(t_i)} \quad \cdot \underbrace{\left( \frac{\kappa}{\| \mathbf{z}_i \|_2} \right).}_{\text{logit-norm awareness }(\gamma_i)}
\]
\reb{Here, $H(x_i)$ is the entropy of the model on $x_i$, $h_0$ is the average entropy on source-domain samples, $\mathbf{z}_i$ is the logit vector, and $\kappa$ is the median logit norm from the source domain.}

\reb{\subsection{Addressing Calibration and Mitigating Pseudo-Label Noise}}

\reb{Our formulation allows us to interpret DEC as implicitly minimizing a calibration-aware surrogate. We named $\tau$ the \textit{certainty regularizer} because it regulates the “sharpness” of predicted probabilities and smoothens the pseudo labels produced by the teacher. As $\tau$ increases, the certainty of the prediction decreases. In other words, $\tau$ should be greater for less reliable observations. This $\tau_i$ is used to smooth teacher predictions via temperature-scaled softmax, and the loss becomes:}
\[
L_{\text{DEC}} = -\sum_{y \in \mathcal{C}} s_i^{\text{teacher}}(y) \log s_i^{\text{student}}(y),
\]
where $s_i^{\text{teacher}} = \texttt{Softmax}(f_{te}(x_i) / \tau_i)$ and $s_i^{\text{student}} = \texttt{Softmax}(f_s(x_i))$.

\reb{This design has three key calibration properties:}
\begin{enumerate}
    \item \textbf{Entropy contrast:} The term $H(x_i) - h_0$ gauges how surprising a prediction is compared to source-domain behavior.
    \item \textbf{Logit-norm awareness:} Smaller $\|\mathbf{z}_i\|$ values yield higher $\tau_i$, adding smoothing when confidence is low in logit space.
    \item \textbf{Sample-wise label smoothing:} The softened teacher predictions act as a regularizer against overconfident or noisy pseudo-labels.
\end{enumerate}

\reb{Thus, DEC behaves like an adaptive version of label smoothing, where each pseudo-label is calibrated on-the-fly based on its reliability. This implicit regularization helps discourage overconfident mispredictions and explains the empirical gains we observe in ECE and NLL.}

\reb{To improve calibration after adaptation, we apply a final temperature scaling step:} $\texttt{Temperature\_Scale}(f_s', \mathtt{avg}(T_{vec}))$. \reb{This temperature is computed from unlabeled test data, preserving the unsupervised setting. It adjusts the student’s logits uniformly, smoothing predictions based on the average certainty regularizer. Because $\tau$ guides the temperature scaling component of our algorithm, we follow the example of \cite{guo2017calibration} and set $t_{min/max}$ within $[1.0, 3.0]$. We recommend setting both these parameters to be higher if the expected domain shift is greater.} To show how DEC regularizes observations appropriately, we continue from Example~\ref{example: no logit norm} to Example~\ref{example: with_logit_norm}:

\begin{example} \label{example: with_logit_norm}

Given the same $g_{t1}, g_{t2}$ and $g_{s}$ from Example~\ref{example: no logit norm}, we input these into our $\texttt{CCR}$ algorithm. We set $h_0 = \mathtt{Entropy_2}(\mathtt{SoftMax}(g_s)), \kappa = |g_s|_2, t_{min} = 1.0$ and $t_{max} = 2.0$. Our algorithm first computes a scaled entropy, $H_{scaled}$ with respect to the source domain entropy for $g_s, g_{t1}$ and $g_{t2}$, as 0.49, 0.48, and 0.50, respectively.

In step 7 of $\texttt{CCR}$, this entropy is transformed into a preliminary regularizer, $t_i$. Then, step 9 adjusts $t_i$ by considering logit norm with respect to the source-domain logit norm. 
\begin{align*}
\mathtt{\tau}_{g_{s}} = 1.49; \quad \mathtt{\tau}_{g_{t2}} = 7.42; \quad \mathtt{\tau}_{g_{t1}} = 1.83
\end{align*}

\end{example}

Notice that, unlike purely entropy-based methods, the $\texttt{CCR}$ algorithm correctly assigns greater regularization to the less reliable samples. Namely, step 9 ensures that samples that are low-entropy due to degenerate reasons are properly regularized by considering logit norm. Furthermore, unique from existing algorithms, our regularizer directly addresses model certainty. The impact of $\texttt{CCR}$ is analyzed in Fig.~\ref{fig:ablation}.

An interpretation as to how our model improves accuracy is through the works of \cite{noisy_student} and \cite{label_smooth_label_noise}. Although the former's work concerns itself in the semi-supervised learning setting, we found their observations to be relevant. That is, they introduce the \textit{noisy student,} a network that has been \textit{noised} by dropout and stochastic-depth. They find that their noisy student can even learn to outperform the teacher which initially produced the pseudo-labels. For the latter work, they establish that label smoothing mitigates label noise, which is a desirable property with respect to unsupervised adaptation. Specifically, they find that label smoothing can be thought of as a regularizer. {This motivates us to smooth more aggressively when we suspect that an observation might be less reliable.}

%% file: tables_charts/HO_Table_mobileNet.tex
\begin{table*}[t]
\LARGE
\centering
\caption{Comparison of Shannon Entropy, ECE, and NLL across the Home Office dataset domains. Results use the MobileNet$_{}$ backbone. {Our reduction in ECE and NLL is statistically significant ($p<0.01$)} while maintaining competitive accuracy.}\label{tab:HO_metrics}
\resizebox{\textwidth}{!}{
\begin{tabular}{l|cccc|cccc|cccc}
\toprule
\textbf{Algorithm} 
& \multicolumn{4}{c|}{\textbf{Shannon Entropy}} 
& \multicolumn{4}{c|}{\textbf{ECE (lower is better)}} 
& \multicolumn{4}{c}{\textbf{NLL (lower is better)}} \\ 
\cmidrule(lr){2-5} \cmidrule(lr){6-9} \cmidrule(lr){10-13}
& \textbf{Art} & \textbf{Clipart} & \textbf{Product} & \textbf{Real World} 
& \textbf{Art} & \textbf{Clipart} & \textbf{Product} & \textbf{Real World} 
& \textbf{Art} & \textbf{Clipart} & \textbf{Product} & \textbf{Real World} \\
\midrule
No Adapt       & 0.950 & 0.951 & 0.704 & 0.689 
               & 0.302 & 0.316 & 0.183 & 0.169 
               & 3.196 & 3.330 & 1.777 & 1.638 \\ 
DEC (ours)     & 2.239 & 2.615 & 2.081 & 1.795 
               & \textbf{0.080} & \textbf{0.047} & \textbf{0.039} & \textbf{0.018} 
               & \textbf{2.077} & \textbf{2.127} & \textbf{1.328} & \textbf{1.212} \\ 
T3A            & 0.188 & 0.231 & 0.138 & 0.170 
               & 0.439 & 0.418 & 0.236 & 0.248 
               & 7.904 & 7.756 & 3.536 & 3.321 \\ 
ETA            & 0.940 & 0.966 & 0.735 & 0.678 
               & 0.313 & 0.324 & 0.208 & 0.172 
               & 3.180 & 3.438 & 1.966 & 1.768 \\ 
TENT           & 0.918 & 0.895 & 0.653 & 0.636 
               & 0.299 & 0.308 & 0.187 & 0.172 
               & 3.120 & 3.234 & 1.780 & 1.632 \\ 
SoTTA          & 0.977 & 0.971 & 0.701 & 0.685 
               & 0.279 & 0.301 & 0.186 & 0.167 
               & 2.938 & 3.240 & 1.825 & 1.632 \\ 
\bottomrule
\end{tabular}
}
\end{table*}

%% file: exp.tex
\section{Experiments} \label{sec:experiments}

\subsection{Experimental Setting}
In order to evaluate DEC, we conduct a series of experiments using three different backbone models across four datasets. Our primary evaluation metrics will be model accuracy, NLL and ECE$_{bins = 15}$ on the observations, allowing us to examine both the predictive performance and the calibration quality of the models. By using varied domains and different backbone architectures, we aim to demonstrate the robustness and adaptability of our algorithm in handling diverse and challenging TTA scenarios. All experiments are run three times. Note that the $t_{min/max}$ hyperparameters are set per dataset; we do not set these per domain shift. {We present accuracy and calibration for all datasets along with experiment variances either in the main paper or Appendix~\ref{sec:appendix_additional_experiments} \& \ref{sec:appendix_r2r_variance}.}

We compare with TENT, T3A, SoTTA and ETA, four recent TTA algorithms which follow the test-time entropy reduction paradigm. We do a single iteration of adaptation for all algorithms unless stated otherwise. Dataset preprocessing steps and {a discussion of hyperparameters for all algorithms are in more detail in the Appendix~\ref{sec:appendix_hyperparams}.}

\subsection{Datasets}
The following publicly available TTA datasets are used in our experiments; we selected these because they are commonly used in existing works and provide a variety of domain shifts. In total, we evaluate our algorithm over {26 domain shifts which include a total of 282 classes.} Furthermore, 15 of our domain shifts have 5 corruption levels. For some datasets, we tested using the ``leave one out'' (LOO) paradigm; for example, in PACS, to test generalization to \textit{pictures}, we first trained our backbone networks on \textit{art, cartoon, sketch} before adapting.

\begin{itemize}[leftmargin=*]
    \item[1.] PACS \cite{PACS} has 4 domains: \textit{pictures, art, cartoon, sketch} with 7 classes. Tested using LOO.
    
    \item[2.] HomeOffice \cite{homeOffice} has 4 domains: \textit{art, clipart, product, real} with 65 classes. Tested using LOO.
    
    \item[3.] Digits is a combination of 3 ``numbers'' datasets: USPS \cite{USPS}, MNIST \cite{mnist}, and SVHN \cite{SVHN}. There are 10 classes. Tested using LOO by training on the source domains' training sets and adapting to target domain's test set. 
    
    \item[4.] TinyImageNet-C (TIN-C) \cite{TIN}, has 15 domains with 200 classes. Backbones are trained on corruption-free (source) training set, adapted to and evaluated on corrupted (target) domains. For each target domain, there are 5 tiers of corruption. 
\end{itemize}

\subsection{Back Bones and Training Details}

We test all but the Digits dataset on two popular classifiers, EfficientNetB0 \cite{effNet} and MobileNet \cite{mobileNet} pre-trained for ImageNet \cite{imageNet}. We flatten the output of both networks and add a final dense layer with an output shape equivalent to the number of classes.

We evaluate the Digits dataset using ``SmallCNN'', which is a minor adaptation of the original LeNet \cite{LeNet} to include batch normalization layers and max pooling. This serves to represent more compact and straightforward architectures for less complex datasets. The specific details and orderings of the layers in SmallCNN are elaborated on in Appendix~\ref{sec:appendix_smallcnn}. Note that all three models use batch normalization layers as necessitated by ETA, TENT, and SoTTA. 





%% file: results.tex
\section{Results}

We present our accuracy, ECE and NLL measurements on the four aforementioned datasets. To show evidence of the over-certainty phenomenon, we also report prediction entropy. More comprehensive figures/tables can be found in the appendix. To show the impact of our $\texttt{CCR}$ algorithm, which produces our certainty regularizer $\tau$, we perform an ablation experiment in Fig.~\ref{fig:ablation}.

\input{tables_charts/tin-c-eff}

\subsection{DEC Reduces Calibration Error}

Due to our algorithm addressing the over-certainty phenomenon, we significantly improve calibration performance. DEC achieves state-of-the-art average ECE and NLL in all tested datasets and in nearly all individual domain shifts. We recognize that reducing entropy \emph{did} improve calibration compared to baseline in some cases, but the resulting calibration was still sub-optimal. Fig.~\ref{fig:ablation} empirically validates our finding that an adaptive certainty regularizer aids in reducing ECE and NLL. Moreover, the variance of $\tau$ increases as the corruption intensity increases; indicating a broader dynamic range of regularization when encountering more difficult observations.

\subsection{DEC Augments Accuracy}
In addition to strong calibration performance, DEC provides consistent accuracy uplifts while not necessitating any transformations on observations. By jointly exploiting backbone entropy and logit norm (see Fig.~\ref{fig:logitNorm}), we are able to effectively assess observation reliability. After doing so, we apply greater regularization to less reliable observations. This, in turn, allows us to mitigate the potential label noise produced by the pseudo-labels.

Tables~\ref{tab:mobilenet_accuracy}, \ref{tab::numbers_avg} and \ref{tab:tin_eff_accuracy} show that our algorithm maintains competitive accuracy with recent test-time entropy minimization approaches while addressing the over-certainty phenomenon. Moreover, unlike SoTTA and T3A, we do not store observations or training samples during the adaptation process as this could potentially cause security or privacy issues during deployment.

\begin{table*}[tbh]
\centering
\caption{Average accuracy, ECE, entropy, and NLL on Digits and PACS datasets tested with LOO. Our approach achieves the best average ECE and NLL. Domain-to-domain $\sigma_{\text{max}}^2$ values for accuracy, ECE, entropy, and NLL are reported.}
\vspace{.1in}
\small
\begin{minipage}{0.45\linewidth}
\centering
\resizebox{0.99\linewidth}{!}{
\begin{tabular}{ccccc}
\toprule
\textbf{Algorithm} & \textbf{Accuracy} & \textbf{ECE} & \textbf{Entropy} & \textbf{NLL} \\ 
\midrule
No Adapt       & 0.589 & 0.301 & 0.439 & 4.465 \\
DEC {\small(ours)} & \textbf{0.648} & \textbf{0.169} & 1.220 & \textbf{1.422} \\
T3A            & 0.625 & 0.271 & 1.873 & 1.889 \\
ETA            & 0.646 & 0.270 & 0.356 & 3.779 \\
TENT           & 0.645 & 0.262 & 0.398 & 3.252 \\
SoTTA          & 0.640 & 0.252 & 0.474 & 3.222 \\
\midrule
$\sigma_{\text{max}}^2$ & 0.220 & 0.180 & 0.200 & 46.240 \\
\bottomrule
\end{tabular}
}
\caption{Performance on the Digits dataset using the SmallCNN$_{}$ backbone.}
\label{tab::numbers_avg}
\end{minipage}
\qquad
\begin{minipage}{0.45\linewidth}
\centering
\resizebox{0.99\linewidth}{!}{
\begin{tabular}{ccccc}
\toprule
\textbf{Algorithm} & \textbf{Accuracy} & \textbf{ECE} & \textbf{Entropy} & \textbf{NLL} \\
\midrule
No Adapt       & 0.873 & 0.105 & 0.081 & 1.182 \\
DEC \small{(ours)} & 0.879 & \textbf{0.061} & 0.233 & \textbf{0.466} \\
T3A            & \textbf{0.897} & 0.084 & 0.065 & 0.866 \\
ETA            & 0.878 & 0.102 & 0.071 & 1.213 \\
TENT           & 0.880 & 0.101 & 0.072 & 1.160 \\
SoTTA          & 0.887 & 0.096 & 0.062 & 1.188 \\
\midrule
$\sigma_{\text{max}}^2$ & 0.020 & 0.014 & 0.075 & 1.848 \\
\bottomrule
\end{tabular}
}
\caption{Performance on the PACS dataset using the EfficientNet$_{}$ backbone. }
\label{tab:pacs_avg_mob}
\end{minipage}

\end{table*}


\begin{table}[h]
\LARGE
\centering
\caption{Accuracy on TIN-C with MobileNet$_{}$ backbone across different tiers of corruption. Standard deviations across the domain shifts are shown as subscripts.}
\label{tab:mobilenet_accuracy}
\resizebox{0.48\textwidth}{!}{
\begin{tabular}{lccccc}
\toprule
\textbf{Algorithm} & \textbf{Tier 1} & \textbf{Tier 2} & \textbf{Tier 3} & \textbf{Tier 4} & \textbf{Tier 5} \\ 
\midrule
No Adapt  & 0.27\textsubscript{0.04} & 0.23\textsubscript{0.05} & 0.19\textsubscript{0.07} & 0.15\textsubscript{0.08} & 0.12\textsubscript{0.08} \\ 
DEC (ours)             & \textbf{0.39}\textsubscript{0.02} & \textbf{0.37}\textsubscript{0.03} & \textbf{0.33}\textsubscript{0.05} & \textbf{0.29}\textsubscript{0.07} & 0.24\textsubscript{0.08} \\ 
T3A            & 0.27\textsubscript{0.04} & 0.24\textsubscript{0.05} & 0.20\textsubscript{0.07} & 0.16\textsubscript{0.08} & 0.13\textsubscript{0.08} \\ 
ETA            & 0.22\textsubscript{0.05} & 0.19\textsubscript{0.06} & 0.14\textsubscript{0.07} & 0.11\textsubscript{0.07} & 0.08\textsubscript{0.06} \\ 
TENT           & 0.37\textsubscript{0.03} & 0.34\textsubscript{0.04} & 0.30\textsubscript{0.06} & 0.25\textsubscript{0.08} & 0.20\textsubscript{0.09} \\ 
SoTTA          & \textbf{0.39}\textsubscript{0.02} & \textbf{0.37}\textsubscript{0.03} & \textbf{0.33}\textsubscript{0.04} & \textbf{0.29}\textsubscript{0.06} & \textbf{0.25}\textsubscript{0.08} \\ 
\bottomrule
\end{tabular}
}

\end{table}

%% file: tables_charts/tin-c-eff.tex

\begin{table*}[h]
\LARGE
\centering
\caption{Comparison of Shannon entropy, ECE, and NLL averaged across the 15 domain shifts of TIN-C. We use the EfficientNet$_{}$ backbone. Standard deviations across domains are shown as subscripts. This experiment highlights how excessive certainty (low Shannon entropy) correlates with sub-optimal calibration. {Our reduction in ECE and NLL is statistically significant ($p<0.01$)} while maintaining competitive accuracy (see Tables~\ref{tab:mobilenet_accuracy} and \ref{tab:tin_eff_accuracy}).}\label{tab:TIN_C_metrics}
\vspace{0.1in}
\resizebox{\textwidth}{!}{
\begin{tabular}{l|ccccc|ccccc|ccccc}
\toprule
\textbf{Algorithm} 
& \multicolumn{5}{c|}{\textbf{Shannon Entropy}} 
& \multicolumn{5}{c|}{\textbf{ECE (lower is better)}} 
& \multicolumn{5}{c}{\textbf{NLL (lower is better)}} \\ 
\cmidrule(lr){2-6} \cmidrule(lr){7-11} \cmidrule(lr){12-16}
& \textbf{Tier 1} & \textbf{Tier 2} & \textbf{Tier 3} & \textbf{Tier 4} & \textbf{Tier 5} 
& \textbf{Tier 1} & \textbf{Tier 2} & \textbf{Tier 3} & \textbf{Tier 4} & \textbf{Tier 5} 
& \textbf{Tier 1} & \textbf{Tier 2} & \textbf{Tier 3} & \textbf{Tier 4} & \textbf{Tier 5} \\
\midrule
No Adapt & 2.00\textsubscript{0.17} & 2.10\textsubscript{0.22} & 2.25\textsubscript{0.30} & 2.44\textsubscript{0.51} & 2.63\textsubscript{0.73} 
         & 0.23\textsubscript{0.02} & 0.26\textsubscript{0.03} & 0.29\textsubscript{0.03} & 0.32\textsubscript{0.04} & 0.34\textsubscript{0.04} 
         & 3.06\textsubscript{0.31} & 3.40\textsubscript{0.45} & 3.92\textsubscript{0.70} & 4.57\textsubscript{0.95} & 5.15\textsubscript{1.07} \\ 
DEC (ours) & 4.48\textsubscript{0.25} & 4.68\textsubscript{0.34} & 4.90\textsubscript{0.53} & 5.01\textsubscript{0.77} & 5.32\textsubscript{0.80} 
           & \textbf{0.08}\textsubscript{0.01} & \textbf{0.08}\textsubscript{0.02} & \textbf{0.07}\textsubscript{0.02} & \textbf{0.06}\textsubscript{0.02} & \textbf{0.04}\textsubscript{0.03} 
           & \textbf{2.27}\textsubscript{0.12} & \textbf{2.44}\textsubscript{0.20} & \textbf{2.69}\textsubscript{0.35} & \textbf{3.04}\textsubscript{0.59} & \textbf{3.38}\textsubscript{0.68} \\ 
T3A      & 1.36\textsubscript{0.11} & 1.41\textsubscript{0.14} & 1.46\textsubscript{0.17} & 1.60\textsubscript{0.20} & 1.74\textsubscript{0.23} 
         & 0.33\textsubscript{0.03} & 0.35\textsubscript{0.03} & 0.38\textsubscript{0.03} & 0.41\textsubscript{0.04} & 0.43\textsubscript{0.04} 
         & 3.63\textsubscript{0.39} & 4.09\textsubscript{0.72} & 5.10\textsubscript{1.80} & 7.13\textsubscript{4.41} & 9.82\textsubscript{7.09} \\ 
ETA      & 1.37\textsubscript{0.08} & 1.47\textsubscript{0.14} & 1.57\textsubscript{0.23} & 1.71\textsubscript{0.38} & 1.87\textsubscript{0.67} 
         & 0.37\textsubscript{0.03} & 0.40\textsubscript{0.03} & 0.41\textsubscript{0.04} & 0.43\textsubscript{0.04} & 0.44\textsubscript{0.04} 
         & 4.45\textsubscript{0.52} & 4.93\textsubscript{0.72} & 4.89\textsubscript{0.64} & 5.63\textsubscript{0.79} & 6.29\textsubscript{1.20} \\ 
TENT     & 1.40\textsubscript{0.08} & 1.48\textsubscript{0.12} & 1.57\textsubscript{0.19} & 1.62\textsubscript{0.33} & 1.79\textsubscript{0.49} 
         & 0.27\textsubscript{0.02} & 0.30\textsubscript{0.03} & 0.33\textsubscript{0.03} & 0.36\textsubscript{0.04} & 0.39\textsubscript{0.04} 
         & 3.01\textsubscript{0.22} & 3.36\textsubscript{0.41} & 3.89\textsubscript{0.75} & 4.54\textsubscript{1.14} & 5.20\textsubscript{1.33} \\ 
SoTTA    & 1.71\textsubscript{0.07} & 1.78\textsubscript{0.12} & 1.88\textsubscript{0.18} & 2.08\textsubscript{0.31} & 2.26\textsubscript{0.41} 
         & 0.20\textsubscript{0.02} & 0.21\textsubscript{0.02} & 0.23\textsubscript{0.03} & 0.25\textsubscript{0.03} & 0.28\textsubscript{0.03} 
         & 2.56\textsubscript{0.13} & 2.75\textsubscript{0.23} & 3.04\textsubscript{0.40} & 3.49\textsubscript{0.75} & 3.98\textsubscript{1.04} \\ 
\bottomrule
\end{tabular}
}
\vspace{0.1in}
\end{table*}

%% file: conclusion.tex
\subsection{Discussion}

We would like to underscore our algorithm's robustness to the choice of $t_{min/max}$. {We performed our experiments by selecting our hyperparameters per dataset, not per domain shift.} As shown in Tables \ref{tab:TIN_C_metrics} and \ref{tab:TIN_C_metrics_mobilenet}, we achieve statistically significant reduction in ECE despite using the same hyperparameters across {15 domain shifts}. Our results in Table~\ref{tab:HO_metrics} further bolster our claims of robustness. Again, our hyperparameters are fixed across the four domain shifts but we still achieve statistically significant improvement. In the interest of reproducibility, we release our code at \url{https://github.com/FinAminToastCrunch/DynamicEntropyControl}.

Our study identifies the \textit{over-certainty phenomenon} of modern TTA methodologies which cause harm to model calibration. To ameliorate this issue, we introduce a certainty regularizer, $\tau$, that modulates model entropy and mitigates pseudo-label noise. The resulting algorithm, DEC, jointly improves model accuracy and reduces calibration error. DEC does not require batch normalization layers like SoTTA, TENT, and ETA do. This permits greater freedom when choosing a backbone. Furthermore, DEC is compatible with existing prototypical learning approaches. \reb{We speculate that there is an additional impact of identifying the over-certainty phenomenon: since existing work relies heavily on entropy-based techniques for assessing reliability, we surmise that improving calibration could engender improved reliability estimates.} A limitation of our work is that we focus only on the classification setting; we make no claim on whether or not the over-certainty phenomenon occurs in regression scenarios.

%% file: appendix.tex

\onecolumn

\section{Appendix}\label{sec: appendix}

\subsection{Experimental Setup Details}

 We used TensorFlow 2.9 \cite{tensorflow2015-whitepaper} with Nvidia CUDNN version 11.3 on an RTX 3080 16GB laptop GPU with 32GB of system memory. All experiments are run three times using $\mathtt{random\_seed = 0, 1, 2}$, respectively.

 \begin{enumerate}
    \item PACS \cite{PACS} has 4 domains: \textit{pictures, art, cartoon, sketch} with 7 classes. All images are resized to $(227,227,3)$ and scaled between $[0,255]$. Tested using LOO. For both MobileNet and EfficientNet: We set $t_{min}$ and $t_{max}$ parameters to $1.00$ and $3.0$ respectively.
    
    \item HomeOffice \cite{homeOffice} has 4 domains: \textit{art, clipart, product, real} with 65 classes. All images are resized to $(128,128,3)$ and scaled between $[0,255]$. Tested using LOO. MobileNet: We set $t_{min}$ and $t_{max}$ parameters to $1.20$ and $2.75$ respectively. EfficientNet: We set $t_{min}$ and $t_{max}$ parameters to $1.00$ and $1.75$ respectively.
    
    \item Digits is a combination of 3 ``numbers'' datasets: USPS \cite{USPS}, MNIST \cite{mnist}, and SVHN \cite{SVHN}. The images are resized to $(32,32,1)$ and scaled between $[0,255]$. There are 10 classes. Tested using LOO by training on the source domains' training sets and adapting to target domain's test set. SmallCNN: We set $t_{min}$ and $t_{max}$ to $1.20$ and $2.75$, respectively.
    
    \item TinyImageNet-C (TIN-C) \cite{TIN},  has 15 domains with 200 classes. All images are resized to $(256,256,3)$ and scaled between $[0,255]$. Backbones are trained on corruption-free (source) training set, adapted to and evaluated on corrupted (target) domains. For each target domain, there are 5 tiers of corruption. MobileNet: We set $t_{min}$ and $t_{max}$ parameters to $1.50$ and $3.00$ respectively. EfficientNet: We set $t_{min}$ and $t_{max}$ parameters to $1.00$ and $2.50$ respectively.  
\end{enumerate}

\subsection{Hyperparameters}\label{sec:appendix_hyperparams}
 We do most initial training on the source domain using $\mathtt{RMS\_Prop(lr = 2e-4)}$ \cite{RMSProp} to minimize cross-entropy loss for $\mathtt{epochs = \{15, 15, 5, 25\}}$ for each enumerated dataset, respectively. SmallCNN is compiled and initially trained with the Adam optimizer \cite{adams_opt} in lieu of RMSProp.  We estimate roughly 1,000 hours of GPU usage at 130 watts of power to conduct our experiments. Note that MobileNet expects inputs to be prepossessed in a unique manner. We use Tensorflow's off-the-shelf pre-processing layer for MobileNet at the input. For the tests of statistical significance, we compared DEC with the second best algorithm for each experiment.\\ 

 \textbf{Note that the following are parameters for what we compare against. Further clarification on their meaning can be found in their respective works}. For ETA, we set $\mathtt{E\_0 = 0.4 \cdot \ln{(C)}}$, as this was their recommended value, and $\mathtt{\epsilon = \{0.6, 0.1, 0.4, 0.125\}}$ for each enumerated dataset, respectively. These $\epsilon$ values were empirically chosen to help their performance. For T3A, we set the number of supports to retain, $M = \infty$, as this provides the lowest calibration error.  For SoTTA, we set $\rho = 0.05$, $\mathcal{C}_0 = \{0.33, 0.33, 0.33, 0.66\}$ for each dataset respectively to help their performance, and $N_{SoTTA} = 64$ as per their recommendations. We use the authors' recommended batch sizes for all techniques. 

 We selected hyper parameters either by the recommendation of the respective authors' or in order to improve performance with respect to calibration error. For example, for ETA we experimented with various values for $\epsilon$ in order to lower their ECE as much as possible per dataset (not per domain shift). A similar process was used for selecting $\mathcal{C}_0$ to aid SoTTA. \\
 
 For our own hyper parameters, we started with $t_{min} = 1.0$ and $t_{max} = 3.0$, then tuned them as we did the other works. We want to emphasize the robustness of our algorithm to the selection of $t_{min/max}$. \textbf{Our hyperparameters are chosen for each dataset, not for individual domain shifts.} As shown in Tables \ref{tab:TIN_C_metrics} and \ref{tab:TIN_C_metrics_mobilenet}, we achieve a statistically significant reduction in ECE and NLL despite using consistent hyperparameters across \textbf{15 domain shifts}. Our results in Table~\ref{tab:HO_metrics} further support our claims of robustness, showing a statistically significant improvement even with fixed hyperparameters across the four domain shifts. Another note is that although step 6 of DEC (Algorithm~\ref{alg:cd_actual_algo}) shows the teacher model inferencing, we can actually just store and reuse the logits computed in line 1 of the \texttt{CCR} Algorithm~\ref{alg:compute_T}. This saves us one forward pass computation. We use a batch size of 50 for our DEC for all experiments. 

 \subsection{Further Discussion of Related Work}\label{sec:appendix_further_discussion_related_work}

 As stated previously, the focus of our work is on \underline{TTA algorithms}. However, we would like to discuss some studies on OOD calibration to provide supplementary background. The authors of \cite{transferableCalibration} and \cite{CCForDomainGeneralization} remark on poor calibration in the OOD setting, however, according to their algorithms, they require the unlabeled-domain-shifted observations \textbf{during training}. That is, their algorithm is not for the deployed setting like ours is. Another work, \cite{onCalibrationAndOOD}, discusses proxies for determining if a model would be calibrated during deployment, before deployment. \reb{Finally, there exist works such as \cite{pseudo-cal} which are post-hoc calibration techniques (not TTA algorithms).} We recognize the relevance of these works but emphasize that they address different facets or settings. 

 Recent work by \cite{entropyEnigma} offers a complementary view on the failure modes of entropy minimization under domain shift. Their study shows that while early adaptation clusters test embeddings near training data and boosts accuracy, continued optimization pushes embeddings away—leading to accuracy collapse. This biphasic behavior helps explain why TTA methods often degrade in calibration over time. While their focus is on representation drift and unsupervised accuracy estimation, our work highlights a parallel failure mode: over-adaptation also induces epistemic miscalibration.

\reb{\subsection{Memory Tradeoffs}}

\reb{The memory overhead of T3A, $\mathrm{P}_{T3A}$, is a function of their hyperparameter $M$ and the size of the final classifier. For SoTTA, the memory overhead is the same as TENT/ETA plus the $N_{SoTTA}$ observations it stores. More formally, by defining $\psi_\omega, \phi_\omega$ as the number of parameters in the feature extractor and classifier, respectively:}
\begin{align}
\small
    \mathrm{P}_{T3A} &= M \cdot \phi_\omega \label{t3a_memory} \\
    \mathrm{P}_{DEC} &= 2\cdot(\psi_\omega + \phi_\omega) \\
    \mathrm{P}_{ETA} = \mathrm{P}_{TENT} &=  \mathtt{Num BN Params} \\
    \mathrm{P}_{SoTTA} &=  \mathtt{Num BN Params } +  N_{SoTTA}\cdot|x|  
\end{align}

\reb{Note that the frozen copy of the backbone (teacher model) used in DEC does not need to be in GPU VRAM at the same time as the backbone being adapted.}

\clearpage
\newpage

\subsection{Additional Experiments}\label{sec:appendix_additional_experiments}

\begin{table*}[!h]
\small
\centering
\caption{Comparison of Shannon entropy, ECE, and NLL averaged across the 15 domain shifts of TIN-C. We use the MobileNet$_{}$ backbone. Standard deviations across domains are shown as subscripts. This experiment highlights how excessive certainty (low Shannon entropy) correlates with sub-optimal calibration. {Our reduction in ECE and NLL is statistically significant ($p<0.01$)} while maintaining competitive accuracy.}
\label{tab:TIN_C_metrics_mobilenet}
\vspace{0.1in}
\resizebox{\textwidth}{!}{
\begin{tabular}{l|ccccc|ccccc|ccccc}
\toprule
\textbf{Algorithm} 
& \multicolumn{5}{c|}{\textbf{Shannon Entropy}} 
& \multicolumn{5}{c|}{\textbf{ECE (lower is better)}} 
& \multicolumn{5}{c}{\textbf{NLL (lower is better)}} \\ 
\cmidrule(lr){2-6} \cmidrule(lr){7-11} \cmidrule(lr){12-16}
& \textbf{Tier 1} & \textbf{Tier 2} & \textbf{Tier 3} & \textbf{Tier 4} & \textbf{Tier 5} 
& \textbf{Tier 1} & \textbf{Tier 2} & \textbf{Tier 3} & \textbf{Tier 4} & \textbf{Tier 5} 
& \textbf{Tier 1} & \textbf{Tier 2} & \textbf{Tier 3} & \textbf{Tier 4} & \textbf{Tier 5} \\
\midrule
No Adapt & 1.44\textsubscript{0.10} & 1.53\textsubscript{0.14} & 1.61\textsubscript{0.18} & 1.67\textsubscript{0.19} & 1.66\textsubscript{0.21} 
         & 0.43\textsubscript{0.02} & 0.44\textsubscript{0.03} & 0.47\textsubscript{0.05} & 0.50\textsubscript{0.06} & 0.53\textsubscript{0.08} 
         & 5.98\textsubscript{0.55} & 6.56\textsubscript{0.88} & 7.36\textsubscript{1.41} & 8.42\textsubscript{2.04} & 9.50\textsubscript{2.66} \\ 
DEC (ours) & 3.35\textsubscript{0.21} & 3.55\textsubscript{0.30} & 3.80\textsubscript{0.42} & 3.97\textsubscript{0.42} & 4.04\textsubscript{0.38} 
           & \textbf{0.05}\textsubscript{0.01} & \textbf{0.05}\textsubscript{0.02} & \textbf{0.06}\textsubscript{0.02} & \textbf{0.07}\textsubscript{0.03} & \textbf{0.10}\textsubscript{0.06} 
           & \textbf{2.68}\textsubscript{0.11} & \textbf{2.83}\textsubscript{0.17} & \textbf{3.05}\textsubscript{0.29} & \textbf{3.37}\textsubscript{0.46} & \textbf{3.78}\textsubscript{0.78} \\ 
T3A      & 1.60\textsubscript{0.10} & 1.70\textsubscript{0.14} & 1.81\textsubscript{0.17} & 1.87\textsubscript{0.17} & 1.89\textsubscript{0.21} 
         & 0.42\textsubscript{0.02} & 0.44\textsubscript{0.02} & 0.46\textsubscript{0.03} & 0.48\textsubscript{0.05} & 0.50\textsubscript{0.06} 
         & 6.36\textsubscript{0.66} & 7.15\textsubscript{1.10} & 9.17\textsubscript{3.37} & 13.27\textsubscript{7.59} & 19.55\textsubscript{15.24} \\ 
ETA      & 1.19\textsubscript{0.10} & 1.22\textsubscript{0.10} & 1.36\textsubscript{0.15} & 1.32\textsubscript{0.19} & 1.34\textsubscript{0.16} 
         & 0.51\textsubscript{0.04} & 0.54\textsubscript{0.05} & 0.56\textsubscript{0.05} & 0.59\textsubscript{0.06} & 0.62\textsubscript{0.06} 
         & 8.69\textsubscript{0.98} & 8.93\textsubscript{0.86} & 9.83\textsubscript{2.17} & 10.68\textsubscript{2.02} & 12.06\textsubscript{2.42} \\ 
TENT     & 1.10\textsubscript{0.04} & 1.14\textsubscript{0.06} & 1.21\textsubscript{0.09} & 1.28\textsubscript{0.12} & 1.34\textsubscript{0.12} 
         & 0.38\textsubscript{0.02} & 0.40\textsubscript{0.03} & 0.43\textsubscript{0.04} & 0.47\textsubscript{0.06} & 0.50\textsubscript{0.07} 
         & 4.95\textsubscript{0.35} & 5.47\textsubscript{0.63} & 6.25\textsubscript{1.17} & 7.35\textsubscript{1.94} & 8.54\textsubscript{2.83} \\ 
SoTTA    & 1.14\textsubscript{0.04} & 1.19\textsubscript{0.05} & 1.26\textsubscript{0.09} & 1.34\textsubscript{0.12} & 1.43\textsubscript{0.16} 
         & 0.36\textsubscript{0.01} & 0.37\textsubscript{0.02} & 0.39\textsubscript{0.03} & 0.42\textsubscript{0.04} & 0.45\textsubscript{0.05} 
         & 4.70\textsubscript{0.23} & 4.99\textsubscript{0.37} & 5.44\textsubscript{0.66} & 6.10\textsubscript{1.01} & 6.89\textsubscript{1.53} \\ 
\bottomrule
\end{tabular}
}
\vspace{0.1in}
\end{table*}

\begin{table}[h]
\small
\centering
\caption{Accuracy on TIN-C with EfficientNet$_{}$ backbone across different tiers of corruption for various algorithms. Standard deviations are shown as subscripts.}\label{tab:tin_eff_accuracy}
\vspace{0.1in}
\resizebox{0.5\textwidth}{!}{
\begin{tabular}{lccccc}
\toprule
\textbf{Algorithm} & \textbf{Tier 1} & \textbf{Tier 2} & \textbf{Tier 3} & \textbf{Tier 4} & \textbf{Tier 5} \\ 
\midrule
No Adapt  & 0.41\textsubscript{0.04} & 0.37\textsubscript{0.06} & 0.31\textsubscript{0.08} & 0.24\textsubscript{0.09} & 0.19\textsubscript{0.10} \\ 
DEC (ours)             & \textbf{0.49}\textsubscript{0.02} & \textbf{0.46}\textsubscript{0.04} & 0.41\textsubscript{0.06} & 0.35\textsubscript{0.10} & 0.29\textsubscript{0.11} \\ 
T3A            & 0.42\textsubscript{0.04} & 0.38\textsubscript{0.06} & 0.32\textsubscript{0.08} & 0.26\textsubscript{0.10} & 0.21\textsubscript{0.10} \\ 
ETA            & 0.33\textsubscript{0.03} & 0.29\textsubscript{0.06} & 0.26\textsubscript{0.07} & 0.22\textsubscript{0.08} & 0.19\textsubscript{0.09} \\ 
TENT           & 0.47\textsubscript{0.03} & 0.42\textsubscript{0.05} & 0.37\textsubscript{0.08} & 0.31\textsubscript{0.10} & 0.25\textsubscript{0.11} \\ 
SoTTA          & \textbf{0.49}\textsubscript{0.02} & \textbf{0.46}\textsubscript{0.03} & \textbf{0.42}\textsubscript{0.06} & \textbf{0.36}\textsubscript{0.09} & \textbf{0.31}\textsubscript{0.10} \\ 
\bottomrule
\end{tabular}
}
\vspace{0.15cm}

\label{tab:accuracy}
\end{table}

\input{tables_charts/HO_table_avg}

\input{tables_charts/Ho_table_avg_mob}
\input{tables_charts/PACS_TABLE_Mob}

\newpage
\subsection{Run-to-Run Variances across all TTA algorithms}\label{sec:appendix_r2r_variance}

\begin{itemize}
\item For the Digits dataset: $\mathtt{\sigma_{max}^2} = [4.39 \times 10^{-3}, 5.51 \times 10^{-3}, 3.85 \times 10^{-3}, 1.00 \times 10^{-3}]$ for accuracy, ECE, and entropy, respectively across all trials using SmallCNN.

\item For Home Office across both backbones: $\mathtt{\sigma_{max}^2} = [5.00 \times 10^{-2}, 1.12 \times 10^{-5}, 1.66 \times 10^{-5}, 7.29 \times 10^{-3}]$ for accuracy, ECE, and entropy, respectively.

\item For the PACS dataset across both backbones: $\mathtt{\sigma_{max}^2} = [1.21 \times 10^{-3}, 1.02 \times 10^{-4}, 2.09 \times 10^{-3}, 2.00 \times 10^{-4}]$ for accuracy, ECE, and entropy, respectively across all trials.

\item For the TIN-C dataset across both backbones and across all 5 domain shifts: $\mathtt{\sigma_{max}^2} = [1.39 \times 10^{-3}, 1.12 \times 10^{-7}, 2.46 \times 10^{-3}, 2.40 \times 10^{-1}]$ for accuracy, ECE, and entropy, respectively across all trials.
\end{itemize}

\clearpage
\subsection{SmallCNN}\label{sec:appendix_smallcnn}
\begin{figure}[!h]
\centering
\includegraphics[width=4.750cm, height=21.0cm]{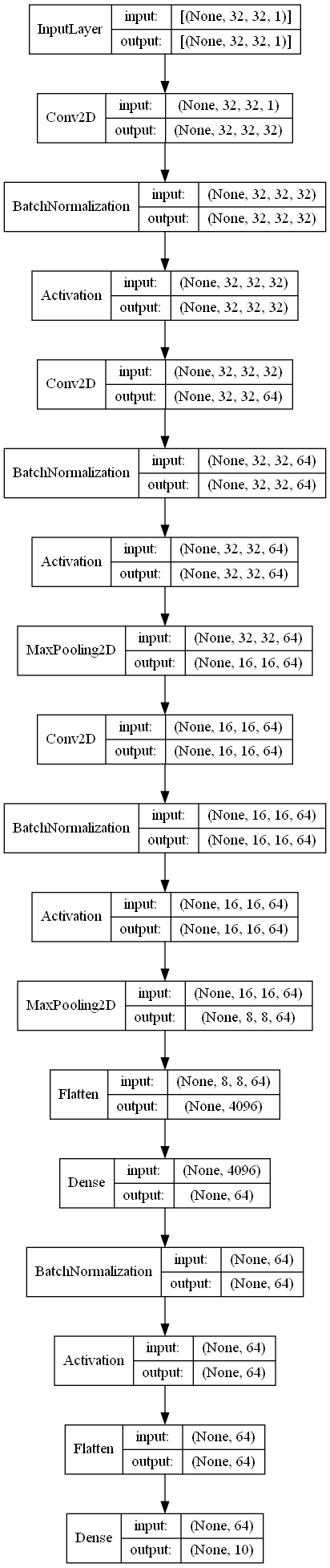} 
\caption{SmallCNN's architecture has 319,498 parameters.} \label{fig:smallCNN}
\end{figure}

%% file: tables_charts/HO_Table_avg.tex

\begin{table}[htp]
\centering
\caption{Average accuracy, ECE, entropy, and NLL on the Home Office dataset using the EfficientNet$_{}$ backbone. Domain-to-domain $\sigma_{\text{max}}^2$ values for accuracy, ECE, entropy, and NLL are reported.}
\label{tab:HO_eff_avg}
\vspace{0.1in}
\resizebox{0.4\textwidth}{!}{
\begin{tabular}{ccccc}
\toprule
\textbf{Algorithm} & \textbf{Accuracy} & \textbf{ECE} & \textbf{Entropy} & \textbf{NLL} \\
\midrule
No Adapt       & 0.665 & 0.198 & 0.642 & 2.184 \\
DEC            & 0.669 & \textbf{0.069} & 1.832 & \textbf{1.490} \\
T3A            & \textbf{0.686} & 0.267 & 0.186 & 4.321 \\
ETA            & 0.665 & 0.218 & 0.422 & 2.745 \\
TENT           & 0.684 & 0.211 & 0.460 & 2.396 \\
SoTTA          & 0.667 & 0.194 & 0.648 & 2.097 \\
\midrule
$\sigma_{\text{max}}^2$ & 0.060 & 0.030 & 1.110 & 3.760 \\
\bottomrule
\end{tabular}
}

\end{table}

%% file: tables_charts/Ho_table_avg_mob.tex

\begin{table}[h]
\centering
\caption{Average accuracy, ECE, entropy, and NLL on the Home Office dataset using the MobileNet$_{}$ backbone. Domain-to-domain $\sigma_{\text{max}}^2$ values for accuracy, ECE, entropy, and NLL are reported.}
\label{tab:HO_avg_mob}
\vspace{0.1in}
\resizebox{0.4\textwidth}{!}{
\begin{tabular}{ccccc}
\toprule
\textbf{Algorithm} & \textbf{Accuracy} & \textbf{ECE} & \textbf{Entropy} & \textbf{NLL} \\
\midrule
No Adapt       & 0.567 & 0.242 & 0.824 & 2.184 \\
DEC            & 0.575 & \textbf{0.046} & 2.183 & \textbf{1.686} \\
T3A            & \textbf{0.617} & 0.335 & 0.182 & 4.321 \\
ETA            & 0.553 & 0.254 & 0.807 & 2.745 \\
TENT           & 0.579 & 0.242 & 0.776 & 2.441 \\
SoTTA          & 0.574 & 0.233 & 0.833 & 2.409 \\
\midrule
$\sigma_{\text{max}}^2$ & 0.013 & 0.011 & 0.116 & 6.467 \\
\bottomrule
\end{tabular}
}

\end{table}


%% file: tables_charts/PACS_TABLE_Mob.tex

\begin{table}[]
\centering
\caption{Average accuracy, ECE, entropy, and NLL on the PACS dataset using the MobileNet$_{}$ backbone. Domain-to-domain $\sigma_{\text{max}}^2$ values for accuracy, ECE, entropy, and NLL are reported.}
\label{tab:pacs_avg_mob}
\vspace{0.1in}
\resizebox{0.4\textwidth}{!}{
\begin{tabular}{ccccc}
\toprule
\textbf{Algorithm} & \textbf{Accuracy} & \textbf{ECE} & \textbf{Entropy} & \textbf{NLL} \\
\midrule
No Adapt       & 0.841 & 0.111 & 0.177 & 1.182 \\
DEC \small{(ours)} & 0.848 & \textbf{0.082} & 0.341 & \textbf{0.466} \\
T3A            & \textbf{0.857} & 0.116 & 0.096 & 0.866 \\
ETA            & 0.842 & 0.101 & 0.192 & 1.213 \\
TENT           & 0.848 & 0.108 & 0.162 & 1.160 \\
SoTTA          & 0.852 & 0.102 & 0.180 & 1.188 \\
\midrule
$\sigma_{\text{max}}^2$ & 0.034 & 0.015 & 0.099 & 1.016 \\
\bottomrule
\end{tabular}
}
\end{table}

%% file: main.bbl
\begin{thebibliography}{61}
\providecommand{\natexlab}[1]{#1}
\providecommand{\url}[1]{\texttt{#1}}
\expandafter\ifx\csname urlstyle\endcsname\relax
  \providecommand{\doi}[1]{doi: #1}\else
  \providecommand{\doi}{doi: \begingroup \urlstyle{rm}\Url}\fi

\bibitem[Abadi et~al.(2015)Abadi, Agarwal, Barham, Brevdo, Chen, Citro, Corrado, Davis, Dean, Devin, Ghemawat, Goodfellow, Harp, Irving, Isard, Jia, Jozefowicz, Kaiser, Kudlur, Levenberg, Man\'{e}, Monga, Moore, Murray, Olah, Schuster, Shlens, Steiner, Sutskever, Talwar, Tucker, Vanhoucke, Vasudevan, Vi\'{e}gas, Vinyals, Warden, Wattenberg, Wicke, Yu, and Zheng]{tensorflow2015-whitepaper}
Mart\'{i}n Abadi, Ashish Agarwal, Paul Barham, Eugene Brevdo, Zhifeng Chen, Craig Citro, Greg~S. Corrado, Andy Davis, Jeffrey Dean, Matthieu Devin, Sanjay Ghemawat, Ian Goodfellow, Andrew Harp, Geoffrey Irving, Michael Isard, Yangqing Jia, Rafal Jozefowicz, Lukasz Kaiser, Manjunath Kudlur, Josh Levenberg, Dandelion Man\'{e}, Rajat Monga, Sherry Moore, Derek Murray, Chris Olah, Mike Schuster, Jonathon Shlens, Benoit Steiner, Ilya Sutskever, Kunal Talwar, Paul Tucker, Vincent Vanhoucke, Vijay Vasudevan, Fernanda Vi\'{e}gas, Oriol Vinyals, Pete Warden, Martin Wattenberg, Martin Wicke, Yuan Yu, and Xiaoqiang Zheng.
\newblock {TensorFlow}: Large-scale machine learning on heterogeneous systems, 2015.
\newblock URL \url{https://www.tensorflow.org/}.
\newblock Software available from tensorflow.org.

\bibitem[Abdar et~al.(2021)Abdar, Pourpanah, Hussain, Rezazadegan, Liu, Ghavamzadeh, Fieguth, Cao, Khosravi, Acharya, Makarenkov, and Nahavandi]{abdar}
Moloud Abdar, Farhad Pourpanah, Sadiq Hussain, Dana Rezazadegan, Li~Liu, Mohammad Ghavamzadeh, Paul Fieguth, Xiaochun Cao, Abbas Khosravi, U.~Rajendra Acharya, Vladimir Makarenkov, and Saeid Nahavandi.
\newblock A review of uncertainty quantification in deep learning: Techniques, applications and challenges.
\newblock \emph{Information Fusion}, 76:\penalty0 243--297, 2021.
\newblock ISSN 1566-2535.
\newblock \doi{https://doi.org/10.1016/j.inffus.2021.05.008}.

\bibitem[Bendale \& Boult(2016)Bendale and Boult]{openMax}
Abhijit Bendale and Terrance~E Boult.
\newblock Towards open set deep networks.
\newblock In \emph{Proceedings of the IEEE conference on computer vision and pattern recognition}, pp.\  1563--1572, 2016.

\bibitem[Creager et~al.(2021)Creager, Jacobsen, and Zemel]{environmentInferenceInvariant}
Elliot Creager, J{\"o}rn-Henrik Jacobsen, and Richard Zemel.
\newblock Environment inference for invariant learning.
\newblock In \emph{International Conference on Machine Learning}, pp.\  2189--2200. PMLR, 2021.

\bibitem[David et~al.(2010)David, Lu, Luu, and P{\'a}l]{impossibility_theorems_for_uda}
Shai~Ben David, Tyler Lu, Teresa Luu, and D{\'a}vid P{\'a}l.
\newblock Impossibility theorems for domain adaptation.
\newblock In \emph{Proceedings of the Thirteenth International Conference on Artificial Intelligence and Statistics}, pp.\  129--136. JMLR Workshop and Conference Proceedings, 2010.

\bibitem[Deng et~al.(2009)Deng, Dong, Socher, Li, Li, and Fei-Fei]{imageNet}
Jia Deng, Wei Dong, Richard Socher, Li-Jia Li, Kai Li, and Li~Fei-Fei.
\newblock Imagenet: A large-scale hierarchical image database.
\newblock In \emph{2009 IEEE conference on computer vision and pattern recognition}, pp.\  248--255. Ieee, 2009.

\bibitem[Fang et~al.(2022)Fang, Li, Lu, Dong, Han, and Liu]{is_ood_learnable}
Zhen Fang, Yixuan Li, Jie Lu, Jiahua Dong, Bo~Han, and Feng Liu.
\newblock Is out-of-distribution detection learnable?
\newblock \emph{Advances in Neural Information Processing Systems}, 35:\penalty0 37199--37213, 2022.

\bibitem[Foret et~al.(2020)Foret, Kleiner, Mobahi, and Neyshabur]{SAM}
Pierre Foret, Ariel Kleiner, Hossein Mobahi, and Behnam Neyshabur.
\newblock Sharpness-aware minimization for efficiently improving generalization.
\newblock \emph{arXiv preprint arXiv:2010.01412}, 2020.

\bibitem[Gong et~al.(2024)Gong, Kim, Lee, Chottananurak, and Lee]{sotta}
Taesik Gong, Yewon Kim, Taeckyung Lee, Sorn Chottananurak, and Sung-Ju Lee.
\newblock Sotta: Robust test-time adaptation on noisy data streams.
\newblock \emph{Advances in Neural Information Processing Systems}, 36, 2024.

\bibitem[Gong et~al.(2021)Gong, Lin, Yao, Dietterich, Divakaran, and Gervasio]{CCForDomainGeneralization}
Yunye Gong, Xiao Lin, Yi~Yao, Thomas~G Dietterich, Ajay Divakaran, and Melinda Gervasio.
\newblock Confidence calibration for domain generalization under covariate shift.
\newblock In \emph{Proceedings of the IEEE/CVF International Conference on Computer Vision}, pp.\  8958--8967, 2021.

\bibitem[Gou et~al.(2021)Gou, Yu, Maybank, and Tao]{kd_survey}
Jianping Gou, Baosheng Yu, Stephen~J Maybank, and Dacheng Tao.
\newblock Knowledge distillation: A survey.
\newblock \emph{International Journal of Computer Vision}, 129:\penalty0 1789--1819, 2021.

\bibitem[Guo et~al.(2017)Guo, Pleiss, Sun, and Weinberger]{guo2017calibration}
Chuan Guo, Geoff Pleiss, Yu~Sun, and Kilian~Q Weinberger.
\newblock On calibration of modern neural networks.
\newblock In \emph{International conference on machine learning}, pp.\  1321--1330. PMLR, 2017.

\bibitem[H{\'e}bert-Johnson et~al.(2018)H{\'e}bert-Johnson, Kim, Reingold, and Rothblum]{multicalibration}
Ursula H{\'e}bert-Johnson, Michael Kim, Omer Reingold, and Guy Rothblum.
\newblock Multicalibration: Calibration for the (computationally-identifiable) masses.
\newblock In \emph{International Conference on Machine Learning}, pp.\  1939--1948. PMLR, 2018.

\bibitem[Hinton et~al.(2015)Hinton, Vinyals, and Dean]{hinton_distil}
Geoffrey Hinton, Oriol Vinyals, and Jeffrey Dean.
\newblock Distilling the knowledge in a neural network.
\newblock In \emph{NIPS Deep Learning and Representation Learning Workshop}, 2015.
\newblock URL \url{http://arxiv.org/abs/1503.02531}.

\bibitem[Howard et~al.(2017)Howard, Zhu, Chen, Kalenichenko, Wang, Weyand, Andreetto, and Adam]{mobileNet}
Andrew~G. Howard, Menglong Zhu, Bo~Chen, Dmitry Kalenichenko, Weijun Wang, Tobias Weyand, Marco Andreetto, and Hartwig Adam.
\newblock Mobilenets: Efficient convolutional neural networks for mobile vision applications, 2017.
\newblock URL \url{https://arxiv.org/abs/1704.04861}.

\bibitem[Hu et~al.(2024)Hu, Liang, Wang, and Foo]{pseudo-cal}
Dapeng Hu, Jian Liang, Xinchao Wang, and Chuan-Sheng Foo.
\newblock Pseudo-calibration: improving predictive uncertainty estimation in unsupervised domain adaptation.
\newblock In \emph{Forty-first International Conference on Machine Learning}, 2024.

\bibitem[Hull(1994)]{USPS}
J.J. Hull.
\newblock A database for handwritten text recognition research.
\newblock \emph{IEEE Transactions on Pattern Analysis and Machine Intelligence}, 16\penalty0 (5):\penalty0 550--554, 1994.
\newblock \doi{10.1109/34.291440}.

\bibitem[Iwasawa \& Matsuo(2021)Iwasawa and Matsuo]{t3a}
Yusuke Iwasawa and Yutaka Matsuo.
\newblock Test-time classifier adjustment module for model-agnostic domain generalization.
\newblock \emph{Advances in Neural Information Processing Systems}, 34:\penalty0 2427--2440, 2021.

\bibitem[Jiang et~al.(2023)Jiang, Liu, Fang, Chen, Liu, Zheng, and Han]{ood_via_prior}
Xue Jiang, Feng Liu, Zhen Fang, Hong Chen, Tongliang Liu, Feng Zheng, and Bo~Han.
\newblock Detecting out-of-distribution data through in-distribution class prior.
\newblock In \emph{International Conference on Machine Learning}, pp.\  15067--15088. PMLR, 2023.

\bibitem[Kingma \& Ba(2014)Kingma and Ba]{adams_opt}
Diederik~P. Kingma and Jimmy Ba.
\newblock Adam: A method for stochastic optimization, 2014.
\newblock URL \url{https://arxiv.org/abs/1412.6980}.

\bibitem[Le \& Yang(2015)Le and Yang]{TIN}
Ya~Le and Xuan Yang.
\newblock Tiny imagenet visual recognition challenge.
\newblock \emph{CS 231N}, 7\penalty0 (7):\penalty0 3, 2015.

\bibitem[LeCun et~al.(1998)LeCun, Bottou, Bengio, and Haffner]{LeNet}
Yann LeCun, L{\'e}on Bottou, Yoshua Bengio, and Patrick Haffner.
\newblock Gradient-based learning applied to document recognition.
\newblock \emph{Proceedings of the IEEE}, 86\penalty0 (11):\penalty0 2278--2324, 1998.

\bibitem[LeCun et~al.(2010)LeCun, Cortes, and Burges]{mnist}
Yann LeCun, Corinna Cortes, and CJ~Burges.
\newblock Mnist handwritten digit database.
\newblock \emph{ATT Labs [Online]. Available: http://yann.lecun.com/exdb/mnist}, 2, 2010.

\bibitem[Lee et~al.(2013)]{psuedo_label}
Dong-Hyun Lee et~al.
\newblock Pseudo-label: The simple and efficient semi-supervised learning method for deep neural networks.
\newblock In \emph{Workshop on challenges in representation learning, ICML}, volume~3, pp.\  896. Atlanta, 2013.

\bibitem[Li et~al.(2017)Li, Yang, Song, and Hospedales]{PACS}
Da~Li, Yongxin Yang, Yi-Zhe Song, and Timothy~M Hospedales.
\newblock Deeper, broader and artier domain generalization.
\newblock In \emph{Proceedings of the IEEE international conference on computer vision}, pp.\  5542--5550, 2017.

\bibitem[Liang et~al.(2020)Liang, Hu, and Feng]{shot}
Jian Liang, Dapeng Hu, and Jiashi Feng.
\newblock Do we really need to access the source data? source hypothesis transfer for unsupervised domain adaptation.
\newblock In \emph{International conference on machine learning}, pp.\  6028--6039. PMLR, 2020.

\bibitem[Liang et~al.(2017)Liang, Li, and Srikant]{liang2017enhancing}
Shiyu Liang, Yixuan Li, and Rayadurgam Srikant.
\newblock Enhancing the reliability of out-of-distribution image detection in neural networks.
\newblock \emph{arXiv preprint arXiv:1706.02690}, 2017.

\bibitem[Lukasik et~al.(2020)Lukasik, Bhojanapalli, Menon, and Kumar]{label_smooth_label_noise}
Michal Lukasik, Srinadh Bhojanapalli, Aditya Menon, and Sanjiv Kumar.
\newblock Does label smoothing mitigate label noise?
\newblock In Hal~Daumé III and Aarti Singh (eds.), \emph{Proceedings of the 37th International Conference on Machine Learning}, volume 119 of \emph{Proceedings of Machine Learning Research}, pp.\  6448--6458. PMLR, 13--18 Jul 2020.
\newblock URL \url{https://proceedings.mlr.press/v119/lukasik20a.html}.

\bibitem[Mancini et~al.(2018)Mancini, Bulo, Caputo, and Ricci]{sourceSpecificNets}
Massimiliano Mancini, Samuel~Rota Bulo, Barbara Caputo, and Elisa Ricci.
\newblock Best sources forward: domain generalization through source-specific nets.
\newblock In \emph{2018 25th IEEE international conference on image processing (ICIP)}, pp.\  1353--1357. IEEE, 2018.

\bibitem[Miao et~al.(2023)Miao, Pang, Li, Bai, and Zheng]{long_tail_ood}
Wenjun Miao, Guansong Pang, Tianqi Li, Xiao Bai, and Jin Zheng.
\newblock Out-of-distribution detection in long-tailed recognition with calibrated outlier class learning.
\newblock \emph{arXiv preprint arXiv:2312.10686}, 2023.

\bibitem[Minderer et~al.(2021)Minderer, Djolonga, Romijnders, Hubis, Zhai, Houlsby, Tran, and Lucic]{revisitingCalibration}
Matthias Minderer, Josip Djolonga, Rob Romijnders, Frances Hubis, Xiaohua Zhai, Neil Houlsby, Dustin Tran, and Mario Lucic.
\newblock Revisiting the calibration of modern neural networks.
\newblock \emph{Advances in Neural Information Processing Systems}, 34:\penalty0 15682--15694, 2021.

\bibitem[Ming et~al.(2022)Ming, Cai, Gu, Sun, Li, and Li]{vl_rep}
Yifei Ming, Ziyang Cai, Jiuxiang Gu, Yiyou Sun, Wei Li, and Yixuan Li.
\newblock Delving into out-of-distribution detection with vision-language representations.
\newblock \emph{Advances in Neural Information Processing Systems}, 35:\penalty0 35087--35102, 2022.

\bibitem[M{\"{u}}ller et~al.(2019)M{\"{u}}ller, Kornblith, and Hinton]{when_does_label_smooth_help}
Rafael M{\"{u}}ller, Simon Kornblith, and Geoffrey~E. Hinton.
\newblock When does label smoothing help?
\newblock \emph{CoRR}, abs/1906.02629, 2019.
\newblock URL \url{http://arxiv.org/abs/1906.02629}.

\bibitem[Netzer et~al.(2011)Netzer, Wang, Coates, Bissacco, Wu, and Ng]{SVHN}
Yuval Netzer, Tao Wang, Adam Coates, Alessandro Bissacco, Bo~Wu, and Andrew~Y Ng.
\newblock Reading digits in natural images with unsupervised feature learning.
\newblock \emph{Neurips Workshop on Deep Learning}, 2011.

\bibitem[Niu et~al.(2022)Niu, Wu, Zhang, Chen, Zheng, Zhao, and Tan]{eata}
Shuaicheng Niu, Jiaxiang Wu, Yifan Zhang, Yaofo Chen, Shijian Zheng, Peilin Zhao, and Mingkui Tan.
\newblock Efficient test-time model adaptation without forgetting.
\newblock In \emph{International conference on machine learning}, pp.\  16888--16905. PMLR, 2022.

\bibitem[Ovadia et~al.(2019)Ovadia, Fertig, Ren, Nado, Sculley, Nowozin, Dillon, Lakshminarayanan, and Snoek]{canyoutrust}
Yaniv Ovadia, Emily Fertig, Jie Ren, Zachary Nado, D~Sculley, Sebastian Nowozin, Joshua~V. Dillon, Balaji Lakshminarayanan, and Jasper Snoek.
\newblock Can you trust your model's uncertainty? evaluating predictive uncertainty under dataset shift, 2019.
\newblock URL \url{https://arxiv.org/abs/1906.02530}.

\bibitem[Pampari \& Ermon(2020)Pampari and Ermon]{pampari2020unsupervised}
Anusri Pampari and Stefano Ermon.
\newblock Unsupervised calibration under covariate shift.
\newblock \emph{arXiv preprint arXiv:2006.16405}, 2020.

\bibitem[Park et~al.(2023)Park, Chai, Yoon, and Teoh]{understand_Feature_norm}
Jaewoo Park, Jacky Chen~Long Chai, Jaeho Yoon, and Andrew Beng~Jin Teoh.
\newblock Understanding the feature norm for out-of-distribution detection.
\newblock In \emph{Proceedings of the IEEE/CVF International Conference on Computer Vision}, pp.\  1557--1567, 2023.

\bibitem[Pleiss et~al.(2017)Pleiss, Raghavan, Wu, Kleinberg, and Weinberger]{onFairnessAndCalibration}
Geoff Pleiss, Manish Raghavan, Felix Wu, Jon Kleinberg, and Kilian~Q Weinberger.
\newblock On fairness and calibration.
\newblock \emph{Advances in neural information processing systems}, 30, 2017.

\bibitem[Press et~al.(2024)Press, Shwartz-Ziv, LeCun, and Bethge]{entropyEnigma}
Ori Press, Ravid Shwartz-Ziv, Yann LeCun, and Matthias Bethge.
\newblock The entropy enigma: Success and failure of entropy minimization.
\newblock \emph{arXiv preprint arXiv:2405.05012}, 2024.

\bibitem[Raina et~al.(2007)Raina, Battle, Lee, Packer, and Ng]{raina2007self}
Rajat Raina, Alexis Battle, Honglak Lee, Benjamin Packer, and Andrew~Y Ng.
\newblock Self-taught learning: transfer learning from unlabeled data.
\newblock In \emph{Proceedings of the 24th international conference on Machine learning}, pp.\  759--766, 2007.

\bibitem[Schlegl et~al.(2017)Schlegl, Seeb{\"{o}}ck, Waldstein, Schmidt{-}Erfurth, and Langs]{AnoGAN}
Thomas Schlegl, Philipp Seeb{\"{o}}ck, Sebastian~M. Waldstein, Ursula Schmidt{-}Erfurth, and Georg Langs.
\newblock Unsupervised anomaly detection with generative adversarial networks to guide marker discovery.
\newblock \emph{CoRR}, abs/1703.05921, 2017.
\newblock URL \url{http://arxiv.org/abs/1703.05921}.

\bibitem[Snell et~al.(2017)Snell, Swersky, and Zemel]{prototype_learning}
Jake Snell, Kevin Swersky, and Richard Zemel.
\newblock Prototypical networks for few-shot learning.
\newblock \emph{Advances in neural information processing systems}, 30, 2017.

\bibitem[Stanton et~al.(2021)Stanton, Izmailov, Kirichenko, Alemi, and Wilson]{does_knowledge_distillation_work}
Samuel Stanton, Pavel Izmailov, Polina Kirichenko, Alexander~A Alemi, and Andrew~G Wilson.
\newblock Does knowledge distillation really work?
\newblock \emph{Advances in Neural Information Processing Systems}, 34:\penalty0 6906--6919, 2021.

\bibitem[Tan \& Le(2019)Tan and Le]{effNet}
Mingxing Tan and Quoc~V. Le.
\newblock Efficientnet: Rethinking model scaling for convolutional neural networks.
\newblock \emph{CoRR}, abs/1905.11946, 2019.
\newblock URL \url{http://arxiv.org/abs/1905.11946}.

\bibitem[Tang et~al.(2020)Tang, Chen, and Jia]{srdc}
Hui Tang, Ke~Chen, and Kui Jia.
\newblock Unsupervised domain adaptation via structurally regularized deep clustering.
\newblock In \emph{Proceedings of the IEEE/CVF conference on computer vision and pattern recognition}, pp.\  8725--8735, 2020.

\bibitem[Teerapittayanon et~al.(2017)Teerapittayanon, McDanel, and Kung]{branchy}
Surat Teerapittayanon, Bradley McDanel, and H.~T. Kung.
\newblock Branchynet: Fast inference via early exiting from deep neural networks.
\newblock \emph{CoRR}, abs/1709.01686, 2017.
\newblock URL \url{http://arxiv.org/abs/1709.01686}.

\bibitem[Tian et~al.(2019)Tian, Zhou, Fan, and Guan]{learning_competitive_descrimintive_re}
Kai Tian, Shuigeng Zhou, Jianping Fan, and Jihong Guan.
\newblock Learning competitive and discriminative reconstructions for anomaly detection.
\newblock In \emph{Proceedings of the AAAI Conference on Artificial Intelligence}, volume~33, pp.\  5167--5174, 2019.

\bibitem[Tieleman et~al.(2012)Tieleman, Hinton, et~al.]{RMSProp}
Tijmen Tieleman, Geoffrey Hinton, et~al.
\newblock Lecture 6.5-rmsprop: Divide the gradient by a running average of its recent magnitude.
\newblock \emph{COURSERA: Neural networks for machine learning}, 4\penalty0 (2):\penalty0 26--31, 2012.

\bibitem[Venkateswara et~al.(2017)Venkateswara, Eusebio, Chakraborty, and Panchanathan]{homeOffice}
Hemanth Venkateswara, Jose Eusebio, Shayok Chakraborty, and Sethuraman Panchanathan.
\newblock Deep hashing network for unsupervised domain adaptation.
\newblock In \emph{Proceedings of the IEEE conference on computer vision and pattern recognition}, pp.\  5018--5027, 2017.

\bibitem[Wald et~al.(2021)Wald, Feder, Greenfeld, and Shalit]{onCalibrationAndOOD}
Yoav Wald, Amir Feder, Daniel Greenfeld, and Uri Shalit.
\newblock On calibration and out-of-domain generalization.
\newblock \emph{Advances in neural information processing systems}, 34:\penalty0 2215--2227, 2021.

\bibitem[Wang et~al.(2020{\natexlab{a}})Wang, Shelhamer, Liu, Olshausen, and Darrell]{TENT}
Dequan Wang, Evan Shelhamer, Shaoteng Liu, Bruno Olshausen, and Trevor Darrell.
\newblock Tent: Fully test-time adaptation by entropy minimization.
\newblock \emph{arXiv preprint arXiv:2006.10726}, 2020{\natexlab{a}}.

\bibitem[Wang et~al.(2022)Wang, Lan, Liu, Ouyang, Qin, Lu, Chen, Zeng, and Yu]{uda_survey}
Jindong Wang, Cuiling Lan, Chang Liu, Yidong Ouyang, Tao Qin, Wang Lu, Yiqiang Chen, Wenjun Zeng, and Philip Yu.
\newblock Generalizing to unseen domains: A survey on domain generalization.
\newblock \emph{IEEE Transactions on Knowledge and Data Engineering}, 2022.

\bibitem[Wang et~al.(2020{\natexlab{b}})Wang, Long, Wang, and Jordan]{transferableCalibration}
Ximei Wang, Mingsheng Long, Jianmin Wang, and Michael Jordan.
\newblock Transferable calibration with lower bias and variance in domain adaptation.
\newblock \emph{Advances in Neural Information Processing Systems}, 33:\penalty0 19212--19223, 2020{\natexlab{b}}.

\bibitem[Wei et~al.(2022)Wei, Xie, Cheng, Feng, An, and Li]{logit_norm}
Hongxin Wei, Renchunzi Xie, Hao Cheng, Lei Feng, Bo~An, and Yixuan Li.
\newblock Mitigating neural network overconfidence with logit normalization.
\newblock In \emph{International Conference on Machine Learning}, pp.\  23631--23644. PMLR, 2022.

\bibitem[Wu et~al.(2023)Wu, Lu, Fang, and Zhang]{meta_ood_detect}
Xinheng Wu, Jie Lu, Zhen Fang, and Guangquan Zhang.
\newblock Meta ood learning for continuously adaptive ood detection.
\newblock In \emph{Proceedings of the IEEE/CVF International Conference on Computer Vision}, pp.\  19353--19364, 2023.

\bibitem[Xie et~al.(2020)Xie, Luong, Hovy, and Le]{noisy_student}
Qizhe Xie, Minh-Thang Luong, Eduard Hovy, and Quoc~V Le.
\newblock Self-training with noisy student improves imagenet classification.
\newblock In \emph{Proceedings of the IEEE/CVF conference on computer vision and pattern recognition}, pp.\  10687--10698, 2020.

\bibitem[You et~al.(2022)You, Phoo, Luo, Zhang, Chao, Hariharan, Campbell, and Weinberger]{UDA_autonomousDriving}
Yurong You, Cheng~Perng Phoo, Katie Luo, Travis Zhang, Wei-Lun Chao, Bharath Hariharan, Mark Campbell, and Kilian~Q Weinberger.
\newblock Unsupervised adaptation from repeated traversals for autonomous driving.
\newblock \emph{Advances in Neural Information Processing Systems}, 35:\penalty0 27716--27729, 2022.

\bibitem[Zenati et~al.(2018)Zenati, Foo, Lecouat, Manek, and Chandrasekhar]{effecient_gan_based_reconstruc}
Houssam Zenati, Chuan~Sheng Foo, Bruno Lecouat, Gaurav Manek, and Vijay~Ramaseshan Chandrasekhar.
\newblock Efficient gan-based anomaly detection, 2018.
\newblock URL \url{https://arxiv.org/abs/1802.06222}.

\bibitem[Zhang et~al.(2021)Zhang, Jiang, Hou, Wei, Han, Li, and Cheng]{delvingdeepLabelSmoothing}
Chang-Bin Zhang, Peng-Tao Jiang, Qibin Hou, Yunchao Wei, Qi~Han, Zhen Li, and Ming-Ming Cheng.
\newblock Delving deep into label smoothing.
\newblock \emph{{IEEE} Transactions on Image Processing}, 30:\penalty0 5984--5996, 2021.
\newblock \doi{10.1109/tip.2021.3089942}.
\newblock URL \url{https://doi.org/10.1109%2Ftip.2021.3089942}.

\bibitem[Zhu et~al.(2022)Zhu, Leibovich, Liu, Mohan, Kaku, Yu, Zanna, Razavian, and Fernandez-Granda]{deepprobability}
Weicheng Zhu, Matan Leibovich, Sheng Liu, Sreyas Mohan, Aakash Kaku, Boyang Yu, Laure Zanna, Narges Razavian, and Carlos Fernandez-Granda.
\newblock Deep probability estimation, 2022.
\newblock URL \url{https://openreview.net/forum?id=hdSn_X7Hfvz}.

\end{thebibliography}
